\documentclass[5p,twocolumn,lefttitle]{elsarticle}

\usepackage{lineno,hyperref}
\modulolinenumbers[5]

\journal{Artificial Intelligence In Medicine}

\usepackage{url}
\usepackage{footmisc}
\usepackage{graphicx}
\usepackage{booktabs}

\usepackage{multirow}
\usepackage{float}

\usepackage{caption}
\usepackage{subcaption}

\usepackage{xcolor}

\usepackage{amsmath}

\usepackage{tabularx}

\begin{document}

\begin{frontmatter}

\title{AIOSA: An Approach to the Automatic Identification of Obstructive Sleep Apnea Events based on Deep Learning\tnoteref{t1}}
\tnotetext[t1]{Final article published on Artificial Intelligence in Medicine: \url{https://www.sciencedirect.com/science/article/pii/S0933365721001263}}

\author[1]{Andrea Bernardini}
\ead{bernardini.andrea@spes.uniud.it}

\author[2]{Andrea Brunello\corref{main}}
\ead{andrea.brunello@uniud.it}

\author[1]{Gian Luigi Gigli}
\ead{gianluigi.gigli@uniud.it}

\author[2]{Angelo Montanari}
\ead{angelo.montanari@uniud.it}

\author[2]{Nicola Saccomanno\corref{main}}
\ead{nicola.saccomanno@uniud.it}

\cortext[main]{Main and corresponding authors.}
\address[1]{Clinical Neurology Unit, Udine University Hospital, Piazzale Santa Maria della Misericordia, 15, 33100 Udine, Italy}
\address[2]{Department of Mathematics, Computer Science, and Physics, University of Udine, Via delle Scienze 206, 33100 Udine, Italy}

\begin{abstract}
Obstructive Sleep Apnea Syndrome (OSAS) is the most common sleep-related breathing disorder. It is caused by an increased upper airway resistance during sleep, which determines episodes of partial or complete interruption of airflow. The detection and treatment of OSAS is particularly important in patients who suffered a stroke, because the presence of severe OSAS is associated with higher mortality, worse neurological deficits, worse functional outcome after rehabilitation, and a higher likelihood of uncontrolled hypertension. The gold standard test for diagnosing OSAS is polysomnography (PSG). Unfortunately, performing a PSG in an electrically hostile environment, like a stroke unit, on neurologically impaired patients is a difficult task; moreover, the number of strokes per day vastly outnumbers the availability of polysomnographs and dedicated healthcare professionals. Hence, a simple and automated recognition system to identify OSAS cases among acute stroke patients, relying on routinely recorded vital signs, is highly desirable. 
The vast majority of the work done so far focuses on data recorded in ideal conditions and highly selected patients, and thus it is hardly exploitable in real-life circumstances, where it would be of actual use. In this paper, we propose a novel convolutional deep learning architecture able to effectively reduce the temporal resolution of raw waveform data, like physiological signals, extracting key features that can be used for further processing. We exploit models based on such an architecture to detect OSAS events in stroke unit recordings obtained from the monitoring of unselected patients. Unlike existing approaches, annotations are performed at one-second granularity, allowing physicians to better interpret the model outcome. 
Results are considered to be satisfactory by the domain experts. Moreover, through tests run on a widely-used public OSAS dataset, we show that the proposed approach outperforms current state-of-the-art solutions.
\end{abstract}


\end{frontmatter}

\section{Introduction}
\label{sec:intro}

Among sleep-related breathing disorders, Obstructive Sleep Apnea Syndrome (OSAS) is the most common one \cite{pmid27568340}. It is caused by an increased upper airway resistance during sleep, leading to episodes of partial or complete interruption of airflow, that bring to phasic reductions in blood oxygen content; arousals from sleep are usually required to interrupt these events. OSAS commonly manifests itself with excessive daytime sleepiness due to sleep fragmentation; however, its most relevant health-related burden is represented by an increased risk of cardiovascular and cerebrovascular accidents such as myocardial infarction and ischemic stroke \cite{pmid24321805}.

\begin{figure*}[t]
    \centering
    \includegraphics[width=0.9\linewidth]{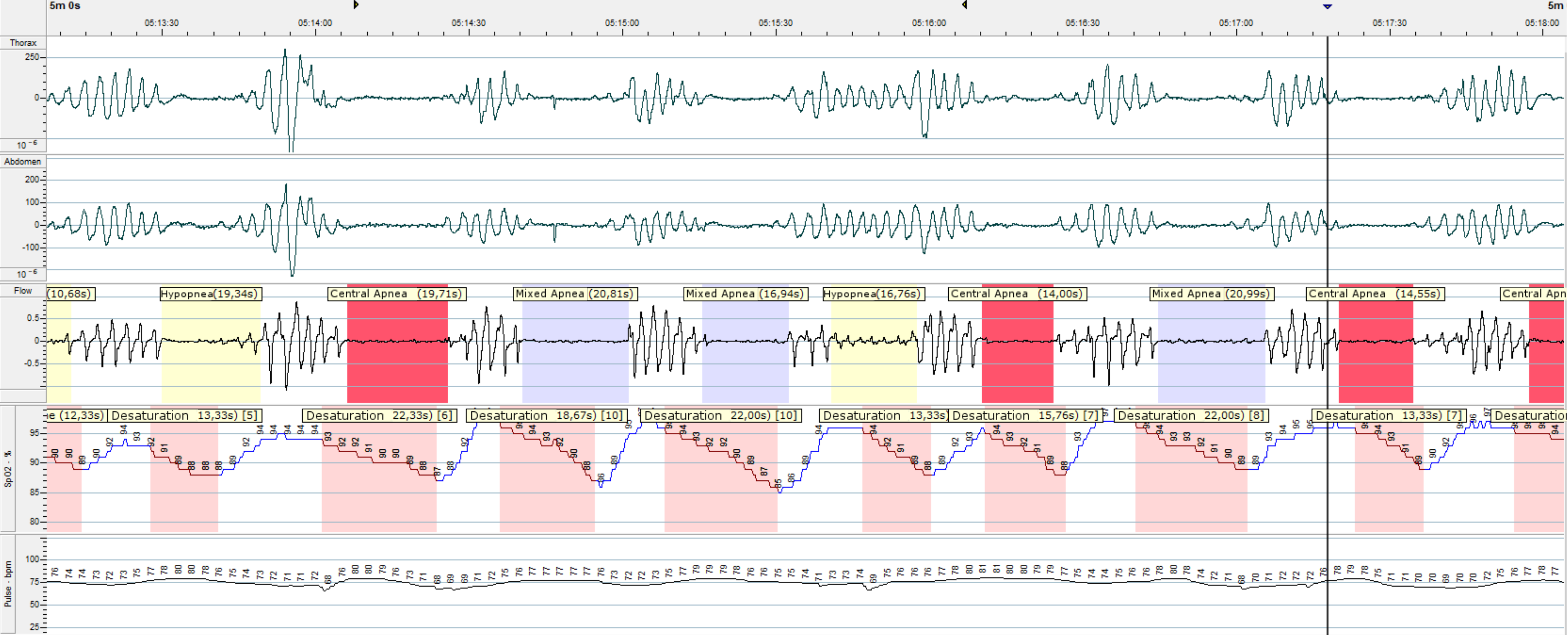}
    \caption{Polysomnographic recording showing some different apnea events.}
    \label{fig:apnea_events}
\end{figure*}

The gold standard test for diagnosing OSAS is polysomnography (PSG), which requires overnight recording of at least the following parameters: airflow, blood oxygen saturation, thoracic and abdominal movements. Moreover, some or all of the following additional parameters are often recorded: snoring, electrocardiography, electroencephalography, electrooculography, surface electromyography of the mylohyoid and tibialis anterior muscles. 
Such recordings are then manually tagged by a trained physician against the presence of apneic events (Figure~\ref{fig:apnea_events}). 
As a result, performing a PSG is labour-, time-, and money-consuming.
Respiratory events can be classified as apneas and hypopneas based on PSG features: the former are characterized by a $>$90\% reduction of respiratory flow for at least 10 seconds, whereas the latter require a $>$30\% reduction of respiratory flow for at least 10 seconds with a concomitant reduction in blood oxygen saturation $\geq$3\% \cite{pmid23066376}. The severity of OSAS is graded by means of a composite measure, named apnea-hypopnea index (AHI), which is calculated dividing the sum of all apneas and hypopneas by the total hours of sleep. OSAS is defined as \emph{mild} when 5$\leq$AHI$<$15, \emph{moderate} when 15$\leq$AHI$<$30, and \emph{severe} when AHI$\geq$30 \cite{PMID:28211654}.

The detection and treatment of OSAS are particularly important in patients who suffered a stroke \cite{pmid29523641}. Stroke is defined as an episode of neurologic dysfunction due to infarction (ischemic stroke) 
or focal collection of blood (hemorrhagic stroke) 
within the central nervous system \cite{pmid23652265}, and represents the second cause of death and the third cause of disability worldwide \cite{GlobalBurdenOfDisease}. The optimal inpatient setting for acute stroke patients is represented by specialized semi-intensive care wards, named stroke units \cite{pmid29367334}. In a stroke unit, all patients undergo continuous monitoring of many vital parameters such as noninvasive blood pressure, multi-lead electrocardiography, photoplethysmography-derived blood oxygen saturimetry, and thoracic impedance-derived respiratory rate.

The prevalence of OSAS is high in the general population, with 49.7\% of men and 23.4\% of women suffering from moderate to severe OSAS \cite{pmid25682233}. In patients with acute stroke, OSAS is even more prevalent, with up to 91.2\% of patients being affected and 44.6\% experiencing severe OSAS \cite{pmid29221777}. 
After an acute stroke, the presence of severe OSAS is associated with higher mortality, worse neurological deficits, worse functional outcome after rehabilitation, and a higher likelihood of uncontrolled hypertension \cite{pmid28522098,pmid25546682}. 

The cornerstone of OSAS treatment is represented by nocturnal continuous positive airway pressure (CPAP) ventilation \cite{pmid16855960}. This noninvasive ventilation system increases air pressure in the upper respiratory tract, thus preventing airway collapse. Treating patients with CPAP determines an improved functional outcome and a reduced 5-year mortality risk due to cardiovascular disease \cite{pmid29523641}.

\begin{figure*}[t]
    \centering
    \includegraphics[width=0.9\linewidth]{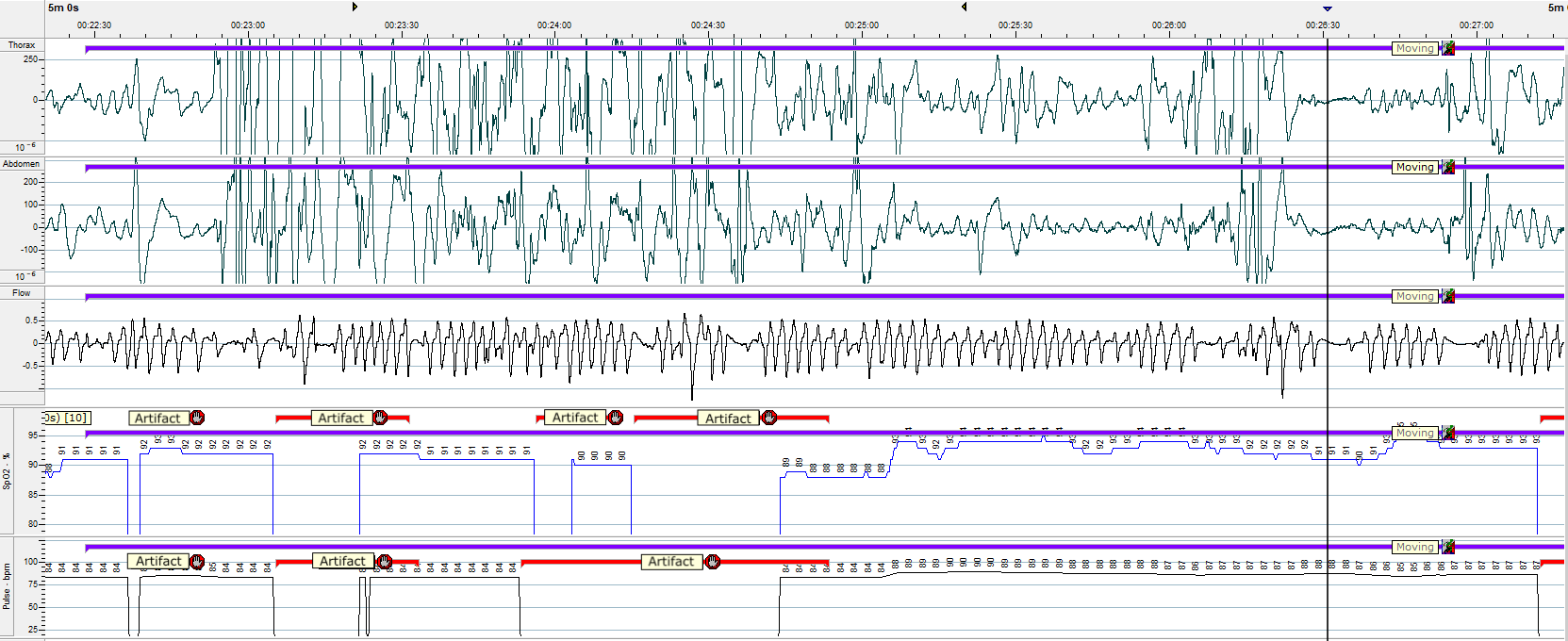}
    \caption{Noisy polysomnographic recording showing different types of artifacts that are common when performing studies in disturbance-prone environments such as a Stroke Unit.}
    \label{fig:artifacts}
\end{figure*}

Unfortunately, performing a PSG in an electrically hostile environment, such as a stroke unit, on neurologically impaired patients is a difficult task, with the result that signals are often affected by noise (Figure \ref{fig:artifacts}); moreover, the number of strokes per day vastly outnumbers the availability of polysomnographs and dedicated healthcare professionals. Therefore, a simple and automated recognition system to identify OSAS cases among acute stroke patients is highly desirable. 

The continuous multiparametric recording of vital signs that is routinely performed in stroke units represents a relevant data source for a comprehensive assessment of patients' health status. However, as stated in \cite{pmid23066376}, diagnosing OSAS with traditional manual sleep scoring requires explicit evaluation of parameters not recorded during stroke unit monitoring, like, for instance, airflow and thoracoabdominal movements. 

Automated analysis of the simplified stroke unit monitoring system may reveal implicit features and thus allow reliable OSAS screening, with no additional procedures or sensors being required and at no extra cost. Similar approaches have been tried before (see Section \ref{sec:related}), with variable success. However, those experiments were performed on data recorded in ideal conditions and on highly selected patients, with stringent exclusion criteria regarding cardiac, respiratory, and other comorbidities. Results obtained under such experimental conditions are hardly generalizable to real-life circumstances, where this solution would be of actual use. In addition, typical approaches are only able to establish whether a patient is affected by OSAS or to roughly locate the presence of anomalous respiratory events during sleep following a coarse-grained windowing or segmentation strategy.

In this paper, we develop a convolutional-based deep learning framework that deals with waveform data by effectively summarizing them and extracting their key properties. Such an architecture can be used as a component in larger models to preprocess raw signals before further elaboration. 
Unlike previous deep learning solutions applied to OSAS detection, the proposed architecture is specifically designed to handle and summarize raw signals with an arbitrarily high sampling frequency, preserving temporal relationships over long time windows. Moreover, to the best of our knowledge, for the first time apnea events are tagged at one-second granularity. Such an ability provides physicians with fine-grained information about the condition of the patient, allowing them to better interpret and validate the results of the model.

We apply the proposed framework to the well-known Apnea-ECG Database \cite{penzel2000apnea}, outperforming current state-of-the-art solutions. Then, we turn to a real case scenario, considering the task of detecting OSAS events during sleep in a stroke unit, with the goal of identifying serious cases. Unlike what happens with existing solutions, the data is collected from the monitoring of unselected patients and include electrocardiogram (ECG) and peripheral oxygen saturation (SpO2). The system is intended to work in an offline fashion, processing overnight recordings as a whole, as typically done in the field. The achieved results are deemed to be satisfactory by domain experts, and may be interpreted as an indicator of the general applicability of the approach in a production setting. This is particularly meaningful as a trained physician necessarily has to rely on the more complex polysomnograph data to perform a similar OSAS assessment. 
The choice of relying on deep learning instead of classical machine learning techniques is motivated by the fact that, as witnessed in the literature \cite{goodfellow2016deep}, deep learning models perform automatic feature extraction. This is of great help since, based on a series of meetings with expert physicians, it emerged that identifying a set of hand-engineered attributes from raw data is quite challenging.

\smallskip

The main contributions of the work are the following:
\begin{itemize}
    \item the design of a novel neural network architecture able to assess OSAS severity and tag apnea events at one-second granularity; 
    \item the network ability to summarize raw physiological signals, reducing their temporal resolution while effectively preserving temporal relationships over long time windows, thanks to the usage of dilated convolutions arranged in a pondered pyramidal scheme;
    \item the validation of the proposed model on a well-established testbed, that confirms its superiority with respect to existing solutions;
    \item the experimental evaluation of the model on a novel Stroke Unit dataset, consisting of data pertaining to unselected patients affected by multiple comorbidities, which suggests the effectiveness of our solution in terms of both classification performance and clinical interpretability of the model 
    output.
\end{itemize}

The paper is organized as follows. Section \ref{sec:related} presents the state of the art in the automatic detection of sleep apnea events. Section \ref{sec:matmeth} illustrates the considered domains. 
In addition, it describes the architectures of the models and the design of the experiments. Section \ref{sec:res} reports the  results obtained from both the Apnea-ECG Database and our dataset. Conclusions provide an assessment of the  work done, and outline future research directions.

\section{Related Work}
\label{sec:related}

A large number of approaches to the automatic identification of sleep apnea and hypopnea events have been proposed in the literature. 

In Table \ref{tab:sota1} and Table \ref{tab:sota2}, we provide a concise, but comprehensive, account of the approaches that make use of ECG and SpO2 recordings, being the far more used and those that we consider in our work. 
As it is well known in the medical domain \cite{wali2020correlation,ucak2021heart}, information conveyed by such signals is strongly related to the presence of OSAS.

\begin{table*}[!ht]
\centering
\caption{Approaches to OSA detection relying on ECG data.}
\label{tab:sota1}
\resizebox{\textwidth}{!}{%
\begin{tabularx}{2\textwidth}{@{}m{2.5em}p{7em}p{10em}lp{2em}p{12em}p{7em}p{3em}p{17em}X@{}}
\toprule
Source                      & Dataset                                                 & Method                                                         & Ref                                      & \# Paz        & Feature type                                                                                                      & Classifier (best)                   & Pred. Gran.                                & Performances (best)                                                                                                                                                                  & Pros \& Cons                                                                                                                                                                                                                                                                                                                                                                          \\ \midrule
\multirow{20}{*}{ECG}       & \multirow{3}{*}{Proprietary}                            & Feature extraction + Feature selection + DL                    & \cite{DBLP:journals/titb/KhandokerGP09}  & 33            & Time/frequency domain statistics, wavelet based                                                                   & Two stage feedforward NN            & 5 sec                                      & Training CV: SE 0.917, SP 0.989, A 0.985. Test set: A 0.947 (apneas), A 0.798 (hypopneas)                                                                                              & [+] Discriminates among apnea and hypopnea; both leave-one-patient-out and CV evaluation. [$-$] Does not consider raw data; lacks comparison on public datasets                                                                                                                                                                                                                                                                                     \\

                        \cmidrule(l){3-10} 
                            &                                                         & Signal filtering + DL                                          & \cite{DBLP:journals/jms/UrtnasanPJL18}   & 82            & Raw data                                                                                                          & 1D CNN                              & 10 sec                                     & Test set: SE 0.96, P 0.96, F1 0.96                                                                                                                                                         & [+] Works on raw data. [$-$] Lacks comparison on public datasets                                                                                                                                                                                                                                                                                                                                                       \\
                        \cmidrule(l){3-10} 
                            &                                                         & Signal filtering + DL                                          & \cite{DBLP:journals/cmpb/UrtnasanKPJL19} & 86            & Raw data or its spectrogram                                                                                    & CNN or LSTM                         & 10 sec                                     & Test set LSTM: SE 0.960, SP 0.960, A 0.960. Test set 1D CNN: SE 0.960, SP 0.960, A 0.963. Test set 2D CNN: SE  0.920, SP 0.910, A 0.912                                                           & [+] Several kinds of DL architectures are compared. [$-$] \textbf{No information on patients exclusion criteria}; lacks comparison on public datasets      \\
                        \cmidrule(l){3-10} 
                            &                                                         &   Feature extraction + DL                                          & \cite{nasifoglu2021obstructive} & 152            & Spectrogram or scalogram                                                                                    & CNN                  & 30 sec                                     &  SE 0.832, SP 0.823, P 0.829, A 0.823                                       & [+] Popular AlexNet, GoogleNet and ResNet18 CNN architectures are compared. [$-$] Lacks comparison on public datasets; does not consider raw data     \\
                        \cmidrule(l){3-10} 
                            &                                                         &   Signal filtering + Feature extraction + Feature selection + ML                                          & \cite{bozkurt2020detection} & 10            & HRV and QRS                                                                                    & Decision tree, KNN, SVM, Ensembles                  & 10 sec                                     &  CV (Ensemble): SE 0.850, SP 0.810, A 0.833                                       & [+] Leave-one-patient-out CV [$-$] \textbf{No information on patients exclusion criteria}; lacks comparison on public datasets; does not consider raw data     \\

        \cmidrule(l){2-10} 
                            & \multirow{13}{*}{Apnea-ECG}                             & Signal filtering + Feature extraction + ML                     & \cite{DBLP:conf/cinc/SadrC14}            & 70            & Time/frequency domain statistics related to RR intervals and EDR signals                                          & ELM                                 & 60 sec                                     & Test set: SE 0.813, SP 0.917, A 0.877                                                                                                                                                   & [+] Official train/test split enabling full reproducibility and fair comparison. [$-$] Does not consider raw data; coarse granularity apnea tagging                                                                                                                                                                                                                                                                                                    \\
                            \cmidrule(l){3-10} 
                            &                                                         & Feature extraction + Feature selection + ML                    & \cite{DBLP:journals/cbm/SharmaS16}       & 70            & QRS Hermite decomposition and rime/frequency domain statistics of RR intervals                                    & SVM                                 & 60 sec                                     & Test set: SE 0.795, SP 0.884, A 0.838, AUC 0.834                                                                                                                                         & [+] Mutiple classifiers are compared; official train/test split enabling full reproducibility and fair comparison. [$-$] Does not consider raw data; coarse granularity apnea tagging                                                                                                                                                                                                                                                                  \\
                            \cmidrule(l){3-10} 
                            &                                                         & Feature extraction + Feature selection + ML                    & \cite{DBLP:journals/tbe/SongLZCX16}      & 70            & Time/frequency domain and other statistics related to RR intervals and EDR signals                                & HMM + SVM                           & 60 sec                                     & Test set: SE 0.826, SP 0.884, A 0.862, AUC 0.940                                                                                                                                        & [+] Leave-one-out CV for feature selection; multiple classification approaches paired with HMM are considered; the HMM shows to be effective at capturing OSA related temporal dependencies; official train/test split enabling full reproducibility and fair comparison. [$-$] Does not consider raw data; coarse granularity apnea tagging                                                                                                         \\
                            \cmidrule(l){3-10} 
                            &                                                         & Feature extraction + DL                                        & \cite{DBLP:conf/cse/ChengSJKL17}         & 35            & RR intervals                                                                                                      & RNN                                 & 60 sec                                     & A 0.978 (?), no further available data                                                                                                                                                & [$-$] The paper misses relevant information regarding the performance evaluation; does not consider raw data; coarse granularity apnea tagging                                                                                                                                                                                                                                                                                                        \\
                            \cmidrule(l){3-10} 
                            &                                                         & DL                                                             & \cite{dey2018obstructive}                & 35            & Raw data                                                                                                          & CNN                                 & 60 sec                                     & Test set: SE 0.978, SP 0.992, P 0.991, A 0.989                                                                                                                                       & [+] Works on raw data. [$-$] \textbf{Train/test split done in a random fashion, implying that data for the same patient are seen both at training and at test time}; coarse granularity apnea tagging                                                                                                                                                                                                                               \\
                            \cmidrule(l){3-10} 
                            &                                                         & Feature extraction + ML/DL                                     & \cite{li2018method}                      & 70            & Stacked autoencoder encoded representation of the RR intervals                                                    & SVM + NN + HMM                      & 60 sec                                     & Test set: SE 0.889, SP 0.821, A 0.847, AUC 0.869                                                                                                                                        & [+] Unsupervised feature extraction; official train/test split enabling full reproducibility and fair comparison. [$-$] Coarse granularity apnea tagging                                                                                                                                                                                                                                                                 \\
                            \cmidrule(l){3-10} 
                            &                                                         & Signal filtering + Feature extraction + DL                     & \cite{singh2019novel}                    & 70            & Scalogram                                                                                                         & 2D CNN + decision fusion classifier & 60 sec                                     & Test set: SE 0.90, SP 0.838, A 0.862, AUC 0.881                                                                                                                                     & [+] Uses a pretrained CNN; official train/test split enabling full reproducibility and fair comparison. [$-$] \textbf{Specific noisy data are removed from the dataset}; does not consider raw data; coarse granularity apnea tagging                                                                                                                                                                                                                                                                                                         \\
                            \cmidrule(l){3-10} 
                            &                                                         & Signal filtering + DL                                          & \cite{DBLP:journals/sensors/ChangYLL20}  & 70            & Raw data                                                                                                          & 1D CNN                              & 60 sec                                     & Test set: SE 0.811, SP 0.920, A 0.879, AUC 0.935                                                                                                                                         & [+] Works on raw data; official train/test split enabling full reproducibility and fair comparison. [$-$] Coarse granularity apnea tagging                                                                                                                                                                                                                                                                                    \\
                            \cmidrule(l){3-10} 
                            &                                                         & Feature extraction + DL                                        & \cite{almutairi2020detection}            & 35            & RR and QRS                                                                                                        & 1D CNN + LSTM                       & 60 sec                                     & Test set: SE 0.899, SP 0.879, A 0.891, F1 0.914                                                                                                                                      & [$-$] \textbf{Plain 10 fold CV is used, with no patient-based splits}; does not consider raw data; coarse granularity apnea tagging                                                                                                                                                                                                                                                                                                                            \\
                            \cmidrule(l){3-10} 
                            &                                                         & Feature extraction + Feature selection + ML                    & \cite{sharma2020sleep}                   & 70            & Relevant features (PCA) extracted from HRV and EDR signals                                                         & KNN (k=32)                          & 60 sec                                     & CV: SE 0.849, SP 0.882, A 0.875, AUC 0.930                                                                                                                                               & [+] Official train/test split enabling full reproducibility and fair comparison; 10 fold patient-grouped CV is considered; multiple classification models are confronted. [$-$] Does not consider raw data; coarse granularity apnea tagging                                                                                                                                                                                                                                                                                           \\
                            \cmidrule(l){3-10} 
                            &                                                         & Signal filtering + Feature extraction + Feature selection + ML & \cite{zarei2020performance}              & 70            & Features extracted by means of an autoregressive analysis                                                         & Random Forest                       & 60 sec                                     & Test set: SE 0.923, SP 0.949, A 0.939, F1 0.92, AUC 0.99                                                                                                                             & [+]  Official train/test split enabling full reproducibility and fair comparison. [$-$] \textbf{Specific noisy data are removed from the dataset (10\%), biasing the overall performance}; does not consider raw data; coarse granularity apnea tagging                                                                                                                                                                                                          \\
                            \cmidrule(l){3-10} 
                            &                                                         & Signal filtering + Feature extraction + DL                     & \cite{feng2020sleep}                     & 70            & Sparse stacked autoencoder encoded representation of the RR intervals                                             & MetaCost + HMM                      & 60 sec                                     & Test set: SE 0.862, SP 0.844, P 0.772, A 0.851, F1 0.814                                                                                                                                  & [+] Unsupervised feature extraction; official train/test split enabling full reproducibility and fair comparison. [$-$] Authors state that the approach seems to be sensitive to different train/test distributions, underlying diseases, and class imbalance ratios; does not consider raw data; coarse granularity apnea tagging                                                                                                                      \\
                            \cmidrule(l){3-10} 
                            &                                                         & Signal filtering + Feature extraction + DL                     & \cite{shen2021multiscale}                & 70            & RR intervals                                                                                                      & 1D CNN                              & 60 sec                                     & Test set: SE 0.898, SP 0.891, P 0.836, A 0.894, F1 0.866, AUC 0.964                                                                                                                       & [+] Analysis of the features learnt by the CNN; official train/test split enabling full reproducibility and fair comparison. [$-$] Does not consider raw data; coarse granularity apnea tagging                                                                                                                                                                                                                                                           \\
                            
                            \cmidrule(l){3-10} 
                            &                                                         &Feature extraction + DL                     & \cite{10.1371/journal.pone.0250618}                & 70            & Spectrogram and scalogram                                                                                                      & 2D CNN                              & 60 sec                                     & CV: SE 0.923, SP 0.926, P 0.890, A 0.924, F1 0.906
                            
                            & [+]  Multiple approaches are confronted. [$-$] Plain 10 fold CV is used, with no patient-based splits; does not consider raw data; coarse granularity apnea tagging                                                                                                                                                                                                                                                         \\                            \cmidrule(l){3-10} 
                            &                                                         & Signal filtering + Feature extraction + ML                     & \cite{FAAL2021102685}                & 35            & Extracted from ARIMA and exponential generalized autoregressive conditional heteroskedasticity models                                                                                        & K-NN                              & 60 sec                                     & Training set random split: SE 0.766, SP 0.844, A 0.814 
                            
                            & [+] Works on (almost) raw data. [$-$] \textbf{Specific noisy data are removed from the dataset}; \textbf{train/test split done in a random fashion, implying that data for the same patient are seen both at training and at test time}; coarse granularity apnea tagging                                                                                                                                                                                                                                                         \\
            \cmidrule(l){2-10} 
                            & Mix of a proprietary and 2 public datasets (Apnea-ECG, SVUH/UCD) & Feature extraction + Feature selection + ML                    & \cite{DBLP:journals/titb/KhandokerPK09}  & 83            & Wavelet decomposition features (variances) obtained from extracted HRV (RR intervals) and EDR (R-wave amplitudes) & SVM                                 & PB                                & Training leave-one-out CV: A 1.0. Test set: A 0.928                                                                                                                                    & [$-$] \textbf{Patients with a history of cardiovascular diseases were excluded to limit the false negative occurrences}; does not consider raw data; only patient-based tagging                                                                                                                                                                                                                                                       \\
            \cmidrule(l){2-10} 
                            & Apnea-ECG + HuGCDN2014                                  & Signal filtering + Feature extraction + Feature selection + ML & \cite{martin2017heart}                   & 70 + 77       & Time based statistics calculated over RR intervals                                                                & LDA or QDA                          & 60 sec                                     & Test set Apnea-ECG: SE 0.814, SP 0.868, A 0.848, AUC 0.92. Test set HuGCDN2014: SE 0.709, SP 0.855, A 0.820, AUC 0.87                                                                  & [+] Official train/test split enabling full reproducibility and fair comparison; Extensive and systematic comparison of different combinations of the elements composing the algorithm. [$-$] Does not consider raw data; coarse granularity apnea tagging                                                                                                                                                                                               \\
            \cmidrule(l){2-10} 
                            & Apnea-ECG + SVUH/UCD                                    & Feature extraction + DL                                        & \cite{wang2019sleep}                     & 70 + 25       & RR intervals and amplitudes                                                                                      & 1D CNN                              & 60 sec                                     & Test Apnea-ECG: SE 0.831, SP 0.903, A 0.876, AUC 0.95. Test SVUH/UCD: SE 0.266, SP 0.869, A 0.718                                                                                           & [+] Several classifiers are compared; official train/test split on Apnea-ECG enabling full reproducibility and fair comparison; second dataset used to validate the results obtained on Apnea-ECG. [$-$] Does not consider raw data; coarse granularity apnea tagging                                                                                                                                                                           \\
            \cmidrule(l){2-10} 
                            & Apnea-ECG + MIT/BIH + SVUH/UCD                          & Feature extraction + Feature selection + ML                    & \cite{fatimah2020detection}              & 35 + (?) + 25 & Relevant features extracted from fourier intrinsic band functions                                                 & SVM                                 & 60 sec                                     & Test set Apnea-ECG: SE 0.897, SP 0.947, P 0.913, A 0.926, AUC 0.97. Test set MIT/BIH: SE 0.881, SP 0.889, A 0.885, AUC 0.940. Test set SVUH/UCD: SE 0.689, SP 0.876, A 0.804, AUC 0.86 & [+] Many combinations of features and models are considered. [$-$] \textbf{Plain 10 fold CV is used, with no patient-based splits}; does not consider raw data; coarse granularity apnea tagging                                                                                                                                                                                                                                                      \\
\bottomrule \\
\multicolumn{10}{l}{\large SE (sensitivity), SP (specificity), P (precision), A (accuracy), AUC (area under the ROC curve), PB (per-patient classification), (?) denotes unknown information, and \textbf{bold} the most critical issues.}
\end{tabularx}%
}
\end{table*}

\begin{table*}[t]
\centering
\caption{Approaches to OSA detection relying on SpO2 and ECG + SpO2 data.}
\label{tab:sota2}
\resizebox{\textwidth}{!}{%
\begin{tabularx}{2\textwidth}{@{}m{2.5em}p{7em}p{10em}lp{2em}p{12em}p{7em}p{3em}p{17em}X@{}}
\toprule
Source                      & Dataset                                                 & Method                                                         & Ref                                      & \# Paz        & Feature type                                                                                                      & Classifier (best)                   & Pred. Gran.                                & Performances (best)                                                                                                                                                                  & Pros \& Cons                                                                                                                                                                                                                                                                                                                                                                          \\ \midrule
\multirow{8}{*}{SpO2}       & \multirow{5}{*}{Proprietary}                            & Feature extraction + Feature selection + ML                    & \cite{alvarez2012feature}                & 240           & Time/frequency domain statistics, spectral features, nonlinear features                                           & Logistic regression                 & PB                                & Test set: SE 0.906, SP 0.813, A 0.875                                                                                                                                                   & [+] An evolutionary algorithm is used to select the most useful features from a large set of candidates. [$-$] Lacks comparison on public datasets; does not consider raw data; only patient-based tagging                                                                                                                                                                                                                                                                              \\
                    \cmidrule(l){3-10} 
                            &                                                         & Signal filtering + Feature extraction + Feature selection + DL & \cite{morillo2013probabilistic}          & 115           & For each patient, overall values of time/frequency domain, statistical and nonlinear features                     & feedforward NN                      & PB                                & Leave-one-out CV: SE 0.924, SP 0.959, A 0.939, AUC 0.97                                                                                                                                 & [+] Leave-one-out CV is considered. [$-$] Lacks comparison on public datasets; does not consider raw data; only patient-based tagging                                                                                                                                                                                                                                                                                                                                                \\
                    \cmidrule(l){3-10} 
                            &                                                         & Signal filtering + Feature extraction + Feature selection + DL & \cite{DBLP:journals/nca/UcarBBP17}       & 5             & Features regarding the shape of the signal, plus statistical values                                               & Feedforward NN                      & Variable & Test set: SE 0.98, SP 0.96, A 0.971                                                                                                                                                       & [+] Thorough statistically analysis on the significance of the extracted features. [$-$] \textbf{Just a few subjects are considered, rising questions about generalizability}; \textbf{train/test split done in a random fashion, implying that data for the same patient are seen both at training and test time}; lacks comparison on public datasets; does not consider raw data                                \\
                    \cmidrule(l){3-10} 
                            &                                                         & Signal filtering + Feature extraction + ML                     & \cite{morales2017sleep}                  & 79            & Time/frequency domain statistics                                                                                  & K-NN                                & PB                                & SE 0.969, SP 0.786, A 0.937                                                                                                                                                          & [$-$] \textbf{It is assumed that an accurate method to segment sleeping and non-sleeping times is available}; \textbf{the test method is not specified}; lacks comparison on public datasets; does not consider raw data; only patient-based tagging.                                                                                                                                                                                                                                    \\
                    \cmidrule(l){3-10} 
                            &                                                         & Manual pattern extraction                                      & \cite{hwang2017real}                     & 230           & Raw data                                                                                                          & Hand-made algorithm                 & 60 sec                                     & Test set: SE 0.828, SP 0.886, A 0.910, PPV 0.838, NPV 0.899                                                                                                                               & [+] Raw outputs are the detected apnea events (start instant + duration in seconds). [$-$] \textbf{Patients exclusion criteria: presence of periodic limb movements, parasomnia, lung disease, chronic chest wall disease, ischemic heart disease or heart failure, anemia}; the authors developed a hand-made algorithm based on patterns emerged from an inspection of time series training data; lacks comparison on public datasets \\
            \cmidrule(l){2-10} 
                            & \multirow{2}{*}{Apnea-ECG}                              & Feature extraction + DL                                        & \cite{almazaydeh2012neural}              & 8             & Time based statistics                                                                                             & Feedforward NN                      & (?) sec               & Test set: SE 0.875, SP 1.0, A 0.933                                                                                                                                                    & [$-$] \textbf{Train/test split is not performed according to the patients}; \textbf{93 instances are considered, extracted from 8 patient recordings, following unclear criteria}; \textbf{just a few subjects are considered, rising questions about generalizability}; \textbf{train/test split done in a random fashion, implying that data for the same patient are seen both at training and at test time}; does not consider raw data        \\
                    \cmidrule(l){3-10} 
                            &                                                         & Feature extraction + Feature selection + DL                    & \cite{mostafa2017optimization}           & 8             & Time/frequency domain statistics                                                                                  & Feedforward NN                      & 60 sec                                     & Test set: SE 0.965, SP 0.985, A 0.977                                                                                                                                                   & [+] An evolutionary algorithm is used to select the most useful features from a large set of candidates. [$-$] \textbf{Just a few subjects are considered, rising questions about generalizability}; \textbf{specific noisy data are removed from the dataset}; \textbf{train/validation/test split done in a random fashion, implying that data for the same patient are seen both at training and at test time}; does not consider raw data; coarse granularity apnea tagging    \\
            \cmidrule(l){2-10} 
                            & Apnea-ECG + SVUH/UCD                                    & DL                                                             & \cite{mostafa2017spo2}                   & 8 + 25        & Raw data                                                                                                          & Deep belief network                 & 60 sec                                     & CV on Apnea-ECG: SE 0.787, SP 0.959, A 0.976. CV on SVUH/UCD: SE 0.604, SP 0.917, A 0.853                                                                                              & [+] Works on raw data. [$-$] \textbf{Specific noisy data are removed from the datasets}; \textbf{plain 10 fold CV is used, with no patient-based splits}; coarse granularity apnea tagging                                                                                                                                                                                                                                              \\
\midrule 
\multirow{5}{0pt}{ECG + SpO2} & \multirow{2}{*}{Proprietary}                            & Feature extraction + Feature selection + ML                    & \cite{ravelo2015oxygen}                  & 70            & Time/frequency domain statistics, linear and non linear features extracted from RR and SpO2 time series           & LDA                                 & 60 sec                                     & Test set: SE 0.734, SP 0.923, A 0.869, AUC 0.919                                                                                                                                         & [+] Raw data are considered. [$-$] \textbf{Patients devoid of other comorbid sleep disorders and heart diseases, including arrhythmia}; lacks comparison on public datasets; does not consider raw data; coarse granularity apnea tagging                                                                                                                                                                                                                                                             \\
            \cmidrule(l){3-10} 
                              &                                                         & Feature extraction + DL + ML                                   & \cite{tuncer2019deep}                    & 100           & PTT signal, converted to a spectrogram                                                                             & 2D CNN + ML (SVM)                   & PB                                & CV: SP 0.980, P 0.942, A 0.928, F1 0.929                                                                                                                                             & [+] It makes use of well-known VGG-16 and AlexNet deep learning models. [$-$] \textbf{Plain 10 fold CV is used, with no patient-based splits}; lacks comparison on public datasets; does not consider raw data; only patient-based tagging                                                                                                                                                                                                                                                    \\
            \cmidrule(l){2-10} 
                            & Apnea-ECG                                               & Feature extraction + DL                                        & \cite{pathinarupothi2017single}          & 35            & HR and SpO2 time series                                                                                           & LSTM                                & 60 sec                                     & Test set: SE 0.847, P 0.995, A 0.921                                                                                                                                                    & [+] A study on the relationship between apnea events, HR, and SpO2 values is conducted. [$-$] \textbf{Train/validation/test split done in a random fashion, implying that data for the same patient are seen both at training and at test time}; does not consider raw data; coarse granularity apnea tagging                                                                                                                                               \\
            \cmidrule(l){2-10} 
                            & SVUH/UCD                                                & Feature extraction + Feature selection + ML/DL                 & \cite{xie2012real}                       & 25            & Time/frequency domain statistics, spectral features, nonlinear features                                           & Decision tree ensemble              & 60 sec                                     & CV: SE 0.797, SP 0.859, A 0.844                                                                                                                                                       & [$+$] A large set of 150 features is considered; several classification models are compared. [$-$] \textbf{Specific noisy data are removed from the dataset}; \textbf{plain 10 fold CV is used, with no patient-based splits}; does not consider raw data; coarse granularity apnea tagging                                                                                                                                                                         \\
                            \bottomrule \\
\multicolumn{10}{l}{\large SE (sensitivity), SP (specificity), P (precision), A (accuracy), AUC (area under the ROC curve), PB (per-patient classification), (?) denotes unknown information, and \textbf{bold} the most critical issues.}
\end{tabularx}%
}
\end{table*}

For each entry, we report information about the type of analyzed data (ECG, SpO2, or both), the kind of dataset (proprietary or not), the size of the study population, the considered features, the employed predictive models, the granularity of the prediction, the overall performance, and the distinctive features and criticalities.  
Providing a complete account of the state of the art on apnea detection is out of the scope of the present work. For further details, including strategies that make use of other kinds of data, we refer the interested reader to one of the many reviews in the literature \cite{mostafa2019systematic,mendonca2018review,hassan2015comparative,pombo2017classification,gutierrez2021reliability}.

Focusing on the contents of the two tables, besides proprietary datasets, many studies rely on the following, publicly-available repositories: Physionet’s Apnea-ECG Database (35 training + 35 test patients) \cite{penzel2000apnea}, SVUH/UCD St. Vincent's University Hospital / University College Dublin Sleep Apnea Database (25 patients) \cite{heneghan2011st}, HuGCDN2014 Database (77 patients) \cite{huyygg}, and MIT-BIH Polysomnographic Database (18 patients) \cite{ichimaru1999development}.

The vast majority of existing solutions make the prediction based on the ECG signal only  \cite{DBLP:journals/titb/KhandokerGP09,DBLP:journals/jms/UrtnasanPJL18,DBLP:journals/cmpb/UrtnasanKPJL19,nasifoglu2021obstructive,bozkurt2020detection,DBLP:conf/cse/ChengSJKL17,dey2018obstructive,li2018method,DBLP:journals/sensors/ChangYLL20,almutairi2020detection,sharma2020sleep,zarei2020performance,feng2020sleep,shen2021multiscale,10.1371/journal.pone.0250618,FAAL2021102685,DBLP:journals/titb/KhandokerPK09,fatimah2020detection}. Moreover, almost all of them make use of some filtering technique to reduce the noise affecting the recordings. In addition, some of them heavily rely on data pre-processing and feature extraction in order to determine relevant attributes to be used in the apnea event prediction task, which may include the calculation of R-R peak intervals, R-wave amplitudes, ECG-derived respiration signals, and the extraction of wavelet coefficients \cite{DBLP:journals/titb/KhandokerGP09,nasifoglu2021obstructive,bozkurt2020detection,DBLP:conf/cse/ChengSJKL17,sharma2020sleep,shen2021multiscale,10.1371/journal.pone.0250618,DBLP:journals/titb/KhandokerPK09,fatimah2020detection}. Other approaches take raw signals as input and apply to them unsupervised feature extraction or exploit solutions that are able of performing feature extraction, such as deep learning models \cite{DBLP:journals/jms/UrtnasanPJL18,DBLP:journals/cmpb/UrtnasanKPJL19,dey2018obstructive,li2018method,DBLP:journals/sensors/ChangYLL20,almutairi2020detection,zarei2020performance,feng2020sleep,FAAL2021102685}.

Several approaches based on SpO2 data only have also been proposed in the literature \cite{alvarez2012feature,morillo2013probabilistic,DBLP:journals/nca/UcarBBP17,morales2017sleep,hwang2017real,almazaydeh2012neural,mostafa2017optimization,mostafa2017spo2}. 
Among them, Mostafa et al.\ \cite{mostafa2017spo2} directly exploit the SpO2 signal following a deep learning approach, while the other ones extract some features from SpO2 data, that are then fed to a proper classification model. These include time-based measures, such as the oxygen desaturation index, stochastic features, including minimum, maximum, variance, and Kurtosis, and \mbox{(time-)frequency-domain} statistics, based on wavelet transformations. 

Finally, there is a smaller set of techniques that combine ECG and SpO2 data \cite{ravelo2015oxygen,tuncer2019deep,pathinarupothi2017single,xie2012real}. Compared to the previously illustrated solutions, such a combination allows one to exploit information coming from a richer set of information sources, and to better deal with missing data and noisy recordings. Data pre-processing and feature extraction steps are carried out to determine possibly relevant features, that include heart rate, R-R intervals, pulse transit time, and statistical values obtained from the SpO2 signal, including maximum, minimum, variance, correlation coefficient, number of zero crossings, and slope.

\begin{figure*}[tb]
    \centering
    \includegraphics[width=\linewidth]{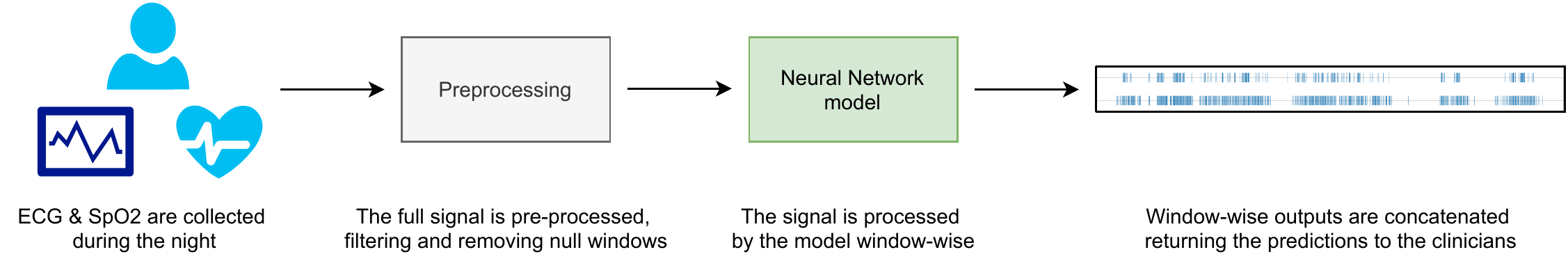}
    \caption{General workflow of AIOSA.}
    \label{fig:AIOSA_flow}
\end{figure*}

Besides the previously analyzed contributions, it is worth mentioning a thorough study of the effectiveness of automatic and human-performed apnea detection conducted by Thorey et al.\ \cite{DBLP:conf/embc/ThoreyHAD19}. 

In such a study, five human scorers are considered and evaluated against each other, and against the performance of a Convolutional Neural Network (CNN). Results show a high variance among the annotations of the different scorers. The performance of the automatic approach is at the level of a sleep expert in the diagnosis of OSAS. 

The quite rich dataset, collected via PSG, includes the following respiratory signals: chest belt, abdominal belt, SpO2, pressure airflow, nasal airflow, and snoring. As for the exclusion criteria, patients with a diagnosed sleep disorder different from obstructive sleep apnea, and individuals suffering from morbid obesity, taking sleep medications, or with complex cardiopulmonary or neurological comorbidities were not considered in the study. 

Results show that the typical human scorer achieves an F1 value of 55\% against the overall consensus, while the machine learning model reaches an F1 score of 57\%, confirming the difficulty of the task.

Except for the last work, all the above contributions exhibit very good performance. 
However, results must be properly weighted as some studies, as it is clearly pointed out in Table \ref{tab:sota1} and Table \ref{tab:sota2}, fragment patients in the experimental evaluation, that is, recordings belonging to a specific subject are split between the training set and the test set, leading to information leakage and raising questions about the generality of the obtained results. In addition, a large portion of the contributions restrict their attention to patients devoid of past or present comorbidities, and, in various cases, preliminarily remove noisy recordings. Last but not least, coarse granularities, up to one minute, are often considered when identifying apnea events, meaning that their exact starting and end points are not identified.

Our setting and goal are quite different. The considered patients suffered from serious strokes and are hospitalized in intensive care units. As a general rule, their clinical situation is considerably complex, and the presence of comorbidities is the rule rather than the exception. In such a context, what matters is a classification of patients based on the presence and the severity of OSAS phenomena. 

To this end, given a patient, we first try to detect and localize all single apnea and hypopnea events at the granularity of one second, and then, exploiting these pieces of information, we derive his/her overall classification. This is quite different from segmenting the patient's sleeping time into intervals of normal and abnormal breathing, thus losing information about the numerosity of OSAS events within each interval, or, even worse, from simply partitioning patients into OSAS or not, providing no information at all about their sleeping conditions, as done by some approaches. 

The analysis of the relevant literature shows that our work significantly differs from existing solutions with respect to the setting, the generated output, and the kind of data taken into consideration.

\section{Materials and Methods}
\label{sec:matmeth}

\begin{table*}[tb]
\centering
\caption{Registration duration before ($\mathcal{T}$) and after ($\widetilde{\mathcal{T}}$) removing all windows with a majority of null values, remaining percentage of null values in ECG and SpO2, AHI value and class, and prediction results under multiple metrics for each patient. }
\label{tab:pats}
\resizebox{\linewidth}{!}{%
\begin{tabular}{c|cccccc|ccccccc}
\toprule
Patient & \begin{tabular}[c]{@{}c@{}}$\mathcal{T}$\\ (hr)\end{tabular} & \begin{tabular}[c]{@{}c@{}}$\widetilde{\mathcal{T}}$\\ (hr)\end{tabular} & \begin{tabular}[c]{@{}c@{}}\% null\\ ECG \end{tabular} & \begin{tabular}[c]{@{}c@{}}\% null\\ SpO2 \end{tabular} &  AHI & \multicolumn{1}{c|}{\begin{tabular}[c]{@{}c@{}}AHI\\ class\end{tabular}} & \multicolumn{1}{c}{\begin{tabular}[c]{@{}c@{}}AHI\\ (pred)\end{tabular}} & \multicolumn{1}{c}{\begin{tabular}[c]{@{}c@{}}AHI class\\ (pred)\end{tabular}} & \multicolumn{1}{c}{Sens} & \multicolumn{1}{c}{Spec} & \multicolumn{1}{c}{Prec} & \multicolumn{1}{c}{Acc} & \multicolumn{1}{c}{F1}  \\ \midrule
1      & 7.0  & 7.0 & 0.0 & 8.3 & 40 & severe   & 41 & severe   & 0.605 & 0.843 & 0.558 & 0.784 & 0.580 \\
2      & 11.9 & 11.9 & 0.0 & 6.7 & 10 & mild     & 21 & moderate & 0.452 & 0.893 & 0.212 & 0.867 & 0.283 \\
3      & 7.1  & 7.1 & 0.2 & 0.2 & 63 & severe   & 64 & severe   & 0.669 & 0.848 & 0.758 & 0.773 & 0.711 \\
4      & 9.0  & 9.0 & 0.1 & 3.3 & 10 & mild     & 5  & mild     & 0.222 & 0.990 & 0.716 & 0.914 & 0.338 \\
5      & 9.1  & 9.0 & 1.8 & 4.1 & 35 & severe   & 18 & moderate & 0.283 & 0.963 & 0.674 & 0.819 & 0.399 \\
6      & 4.1  & 4.0 & 0.4 & 1.6 & 58 & severe   & 30 & severe   & 0.390 & 0.944 & 0.754 & 0.776 & 0.514 \\
7      & 9.2  & 9.2 & 0.0 & 12.5 & 30 & severe   & 21 & moderate & 0.257 & 0.916 & 0.319 & 0.829 & 0.285 \\
8$^*$  & 9.0  & 9.0 & 0.0 & 14.3 & 1  & none     & 3  & none     & 0.170 & 0.989 & 0.102 & 0.983 & 0.127 \\
9$^*$  & 9.5  & 9.4 & 0.3 & 16.9 & 8  & mild     & 12 & mild     & 0.391 & 0.956 & 0.286 & 0.931 & 0.330 \\
10$^*$ & 11.4 & 11.4 & 5.8 & 18.6 & 41 & severe   & 51 & severe   & 0.552 & 0.753 & 0.467 & 0.697 & 0.506 \\
11     & 8.4  & 8.4 & 0.0 & 23.2 & 4  & none     & 4  & none     & 0.238 & 0.982 & 0.215 & 0.967 & 0.226 \\
12     & 8.4  & 8.3 & 0.0 & 2.6 & 4  & none     & 7  & mild     & 0.509 & 0.969 & 0.267 & 0.959 & 0.350 \\
13     & 9.5  & 9.4 & 0.0 & 3.1 & 26 & moderate & 31 & severe   & 0.573 & 0.880 & 0.414 & 0.840 & 0.481 \\
14$^*$ & 10.3 & 10.2 & 0.0 & 8.5 & 9  & mild     & 14 & mild     & 0.487 & 0.919 & 0.335 & 0.886 & 0.397 \\
15     & 9.2  & 9.2 & 0.0 & 19.5 & 43 & severe   & 20 & moderate & 0.312 & 0.954 & 0.778 & 0.737 & 0.446 \\
16     & 9.0  & 9.0 & 12.0 & 13.3 & 37 & severe   & 44 & severe   & 0.761 & 0.864 & 0.773 & 0.825 & 0.767 \\
17$^*$ & 8.1  & 8.0 & 0.0 & 66.3 & 28 & moderate & 33 & severe   & 0.675 & 0.901 & 0.587 & 0.862 & 0.628 \\
18$^*$ & 10.0 & 9.9 & 0.0 & 44.6 & 4  & none     & 1  & none     & 0.0   & 0.996 & 0.0   & 0.981 & 0.0   \\
19     & 9.5  & 9.4 & 0.0 & 29.0 & 10 & mild     & 7  & mild     & 0.225 & 0.971 & 0.278 & 0.935 & 0.249 \\
20$^*$ & 9.8  & 9.8 & 0.0 & 5.2 & 48 & severe   & 35 & severe   & 0.658 & 0.855 & 0.707 & 0.786 & 0.681 \\
21     & 9.8  & 9.7 & 0.0 & 0.3 & 28 & moderate & 31 & severe   & 0.598 & 0.881 & 0.497 & 0.835 & 0.543 \\
22     & 8.0  & 8.0 & 0.0 & 41.3 & 2  & none     & 2  & none     & 0.558 & 0.990 & 0.517 & 0.983 & 0.537 \\
23     & 9.5  & 9.4 & 0.0 & 18.4 & 0  & none     & 0  & none     & 0.0   & 0.998 & 0.0   & 0.996 & 0.0   \\
24     & 9.8  & 9.8 & 0.0 & 14.3 & 21 & moderate & 22 & moderate & 0.521 & 0.927 & 0.571 & 0.863 & 0.544 \\
25     & 10.6 & 10.5 & 0.0 & 18.6 & 44 & severe   & 22 & moderate & 0.304 & 0.918 & 0.525 & 0.778 & 0.385 \\
26     & 9.0  & 9.0 & 0.0 & 12.6 & 60 & severe   & 57 & severe   & 0.706 & 0.819 & 0.697 & 0.777 & 0.701 \\
27     & 7.6  & 7.6 & 0.0 & 19.0 & 9  & mild     & 11 & mild     & 0.307 & 0.953 & 0.263 & 0.920 & 0.283 \\
28     & 7.4  & 7.4 & 0.0 & 73.1 & 13 & mild     & 8  & mild     & 0.180 & 0.969 & 0.377 & 0.893 & 0.243 \\
29     & 7.8  & 7.8 & 0.0 & 13.5 & 4  & none     & 6  & mild     & 0.454 & 0.977 & 0.298 & 0.966 & 0.360 \\
30     & 8.2  & 8.1 & 0.0 & 0.5 & 73 & severe   & 39 & severe   & 0.429 & 0.892 & 0.693 & 0.725 & 0.530 \\
\bottomrule
\multicolumn{13}{l}{\small Patients with an asterisk ($^*$) are those in the validation set.}
\end{tabular}
}
\end{table*}

In this section, we describe the datasets, the developed models,  the experimental setting, and the proposed neural network architecture. 
Overall, our work differs from existing ones in at least three fundamental aspects: \emph{(i)}~we focus on a real-world scenario, where patient exclusion criteria are much less stringent than those usually applied; \emph{(ii)}~we tag apnea and hypopnea events at a 1-second granularity, considerably enhancing the interpretability of the output; and, \emph{(iii)} a distinctive design feature of the proposed architecture is its ability to summarize raw signals with an arbitrarily high sampling frequency, preserving temporal relationships over large time windows. 

A general overview of the workflow of AIOSA is depicted in Figure~\ref{fig:AIOSA_flow}. It is an offline approach, that processes the overnight recordings of a patient as a whole. All experiments have been run on a virtual machine hosted on Google Cloud Platform, equipped with 8 virtual CPUs, 60 GB of RAM, and a v3.8 TPU. As for the development framework, we relied on PyTorch 1.6. For reproducibility purposes, the source code of the developed models will be made available online at \url{https://github.com/dslab-uniud/OSAS}.

\subsection{The Datasets}
\label{sec:domain}

\begin{figure}[t]
    \centering
    \includegraphics[width=\linewidth]{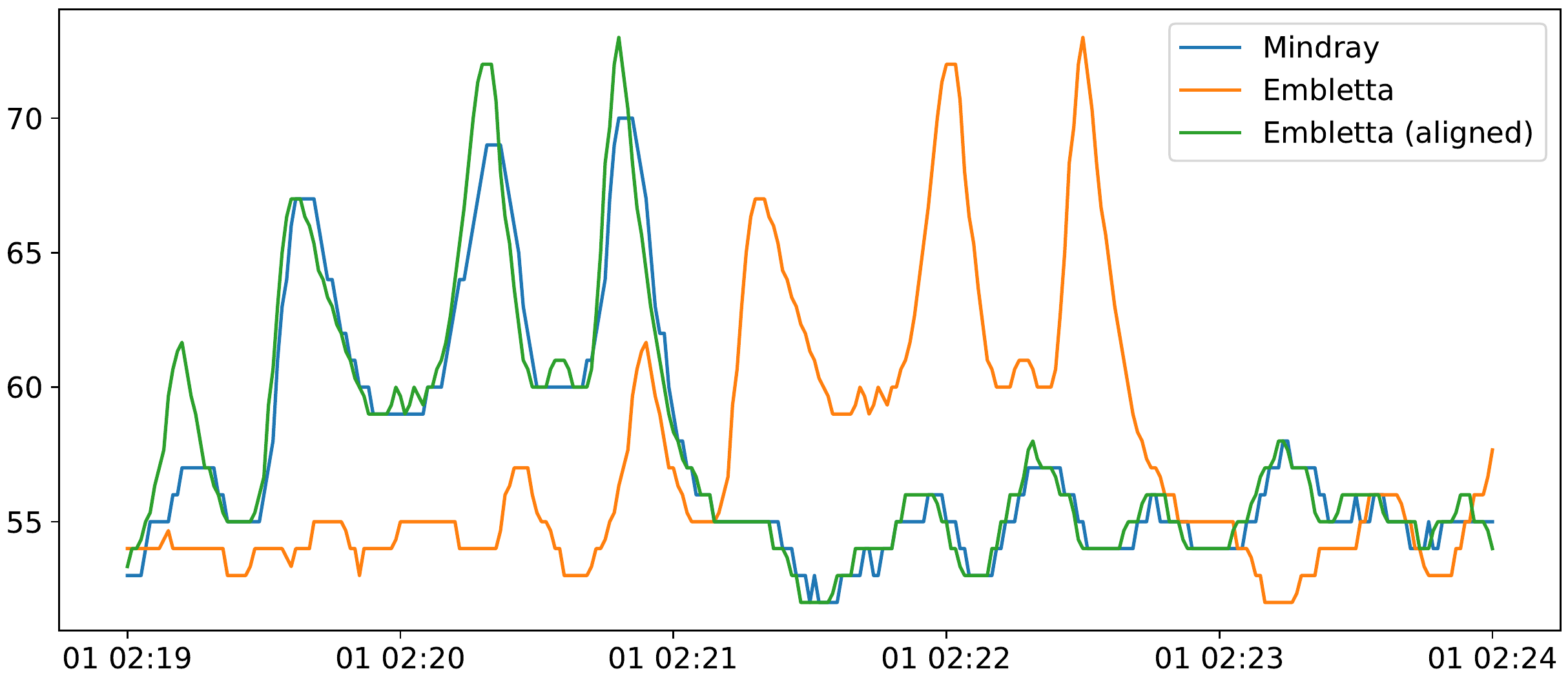}
    \caption{Embletta (original and aligned) and Mindray heart rate signals (5-minute interval).}
    \label{fig:align}
\end{figure}

The first dataset we considered is Physionet's Apnea-ECG Database. It consists of 70 recordings, each belonging to a different subject, equally divided into a training and a test set. Patients are affected by different OSAS severity, including control subjects without OSA (AHI $<$ 5). Each entry consists of a relatively non-noisy and continuous single-lead ECG signal sampled at 100 Hz with 16 bit resolution, and a set of apnea annotations at 1-minute granularity. The duration of the recordings ranges from roughly 7 to 10 hours each. The recordings are segmented into 1 minute intervals, each tagged as either normal or apnea by a human scorer. 

\begin{figure}[t]
    \centering
    \includegraphics[width=1.0\linewidth]{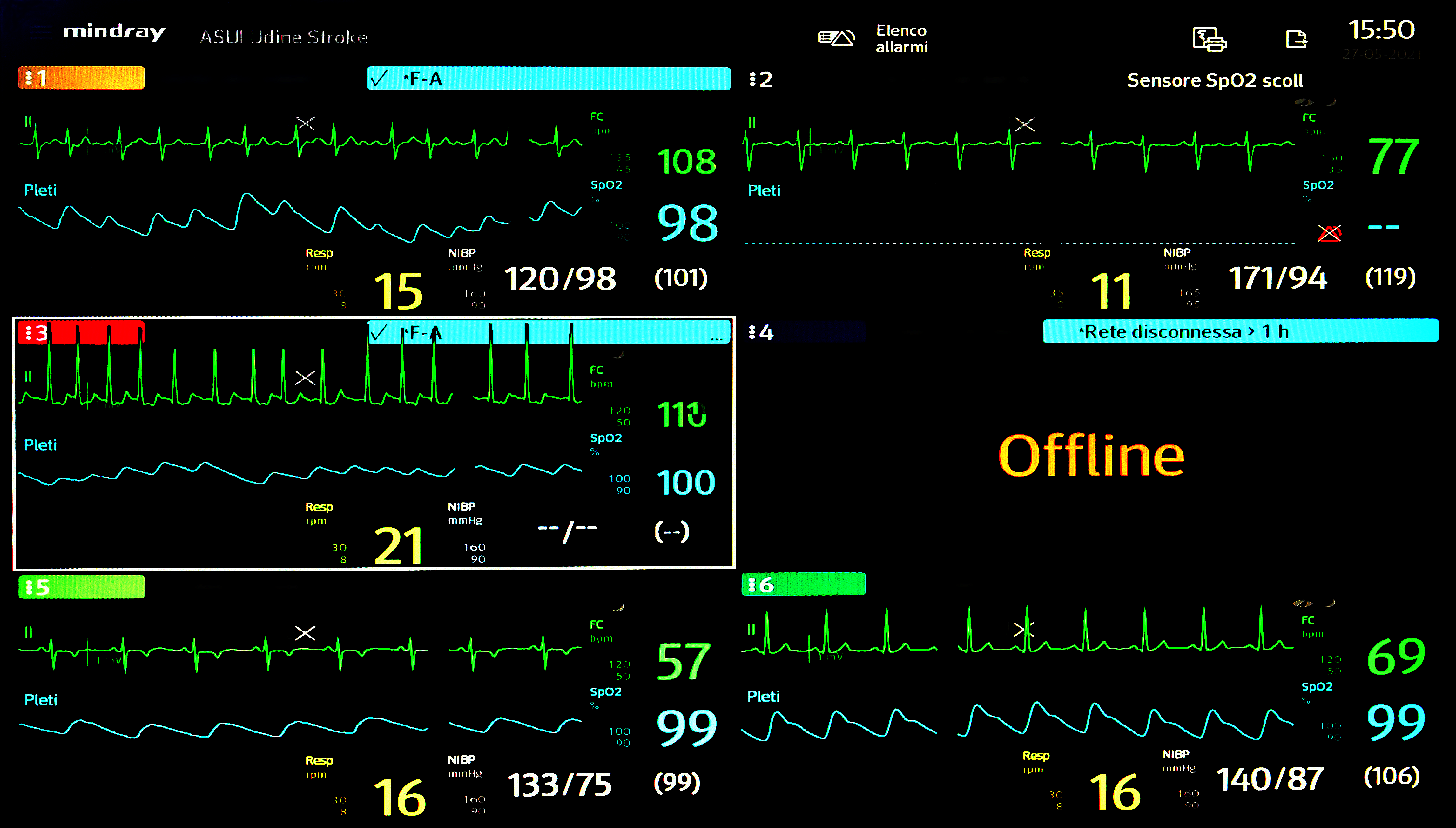}
    \caption{ECG waveform and photoplethysmography-derived SpO2 blood oxygen saturation collected by Mindray Benevision CMS II.}
    \label{fig:mindray}
\end{figure}

As for our dataset, whose summary can be found on the left side of Table~\ref{tab:pats}
, we collected data about 30 patients who were admitted to the stroke unit of the Clinical Neurology Unit of the Udine University Hospital for a suspected cerebrovascular event (ischemic stroke, transient ischemic attack, or hemorrhagic stroke) from August 2019 to July 2020. Patients were screened for the following exclusion criteria: age $<$18 years, insufficient compliance to standard monitoring and/or PSG, aphasia of sufficient severity to limit comprehension of the study protocol and/or expression of informed consent, high risk of alcohol/drug withdrawal syndrome. Diabetes mellitus, atrial fibrillation, cardiac disease, obesity, and other medical conditions not listed above were not considered as exclusion criteria. 
After giving informed consent, patients 
underwent simultaneous overnight vital signs and PSG recording.
Vital signs were collected by a Mindray iMec15 monitor 
connected to a Mindray Benevision CMS II central monitoring system 
and, among them, we considered ECG waveform (II derivation, 80 Hz) and photoplethysmography-derived SpO2 blood oxygen saturation (1 Hz). A graphical account of the recorded data is depicted in Figure~\ref{fig:mindray}.
PSG was performed with an Embletta MPR polysomnograph, 
recording the following channels: thoracic movements, abdominal movements, nasal airflow, blood oxygen saturation, snoring, body position, and movement activity. Recordings were analyzed with Embla RemLogic Software 
by trained sleep medicine physicians in accordance with the American Academy of Sleep Medicine sleep scoring rules \cite{pmid23066376}, and tagged against the presence of central/obstructive/mixed apnea and hypopnea events (which we refer to as anomalies), each identified by its specific time interval.\footnote{We look forward to publishing the dataset, after a proper anonymization phase, and with the consent of the involved authorities.}

\subsection{Data Preprocessing}
\label{sec:data_preproc}

As for the Apnea-ECG Database, preprocessing was performed by applying a Butterworth bandpass filter of order 2, with 5 Hz highpass frequency and 35 Hz lowpass frequency on the ECG waveform signals, as suggested by the literature \cite{mostafa2019systematic}. Then, a 180 seconds-worth of ECG data, that is, the interval of 60 seconds corresponding to the label, and the intervals of 60 seconds respectively preceding and following it, was associated with each binary label, in order to track the general presence or absence of apnea in a specific time interval. Given the 100 Hz ECG sampling rate, we ended up with 18000 predictor values per instance. No instance was discarded from the dataset.

As for our dataset, we performed a more elaborated preprocessing task. 
First, upon inspection of the gathered data, we observed a misalignment between the signals recorded by Embletta and Mindray devices, caused by different settings of their reference clocks. Thus, in order to correctly associate the anomaly intervals (tagged with respect to Embletta) with the Mindray data, for each patient, we looked at the concordance between the heart rate signal recorded by the two devices. Figure~\ref{fig:align} shows the situation for the heart rate signal of one of the considered patients.

Then, we focused on the Mindray data, which are the input to our models, applying again a Butterworth bandpass filter of order 2, with 5 Hz highpass frequency and 35 Hz lowpass frequency on the ECG waveform signals. 
In order to assemble the final dataset, ground truth anomaly data were arranged into 60-second, non-overlapping windows, each encoded by a list of 60 binary values, that keep track of the presence (true) or absence (false) of an anomaly at 1-second granularity. The associated predictors were 180 seconds-worth of ECG and SpO2 data, that is, the 60-second interval corresponding to the labels, and the 60-second intervals respectively preceding and following them. 
All input arrays were independently normalized, mapping them to the interval 0--1 based on their minimum and maximum values. As a result, each instance is characterized by 14400 (180$\cdot$80) ECG values, 180 SpO2 values, and 60 binary labels, all one-dimensional.

It is worth pointing out that, even though the length of the look-back/look-ahead predictors' windows was empirically determined based on training set data, it makes perfect sense from a clinical point of view, as confirmed by the trained physicians involved in the study. Roughly speaking, they allow the neural network to reason about the context that surrounds the 60-second target window. The context contribution is threefold: it improves the detection of deviations from baseline normal breathing behaviour, it makes us aware of other anomalies happening in close proximity, taking into account that apnea events tend to cluster together, and it facilitates the detection of OSA events occurring at the edges of the target window.

As a final step, we reasoned about missing values, whose presence is to be expected given the 
characteristics of the dataset. For instance, monitors may experience malfunctions or sensors may disconnect during the night due to patients' movements. We observed 4\% of null ECG data and 16\% of null SpO2 data. We decided to keep just the instances in the dataset with at least 50\% of non-null SpO2 or ECG values. The resulting recording lengths and null percentages for each patient are reported in Table \ref{tab:pats}. As it can be seen, following such an approach we discarded just a small number of instances, that would not have been useful for model training and assessment, due to their reduced information content. We still kept a consistent amount of null values, replaced by \mbox{-1}, to allow the models to learn how to cope with missing real-world data.

\subsection{Experimental Setting}

\begin{figure}[htb]
\centering
\begin{tabular}[t]{cc}
    \begin{subfigure}{0.5\linewidth}
        \centering
        \includegraphics[width=0.8\linewidth,]{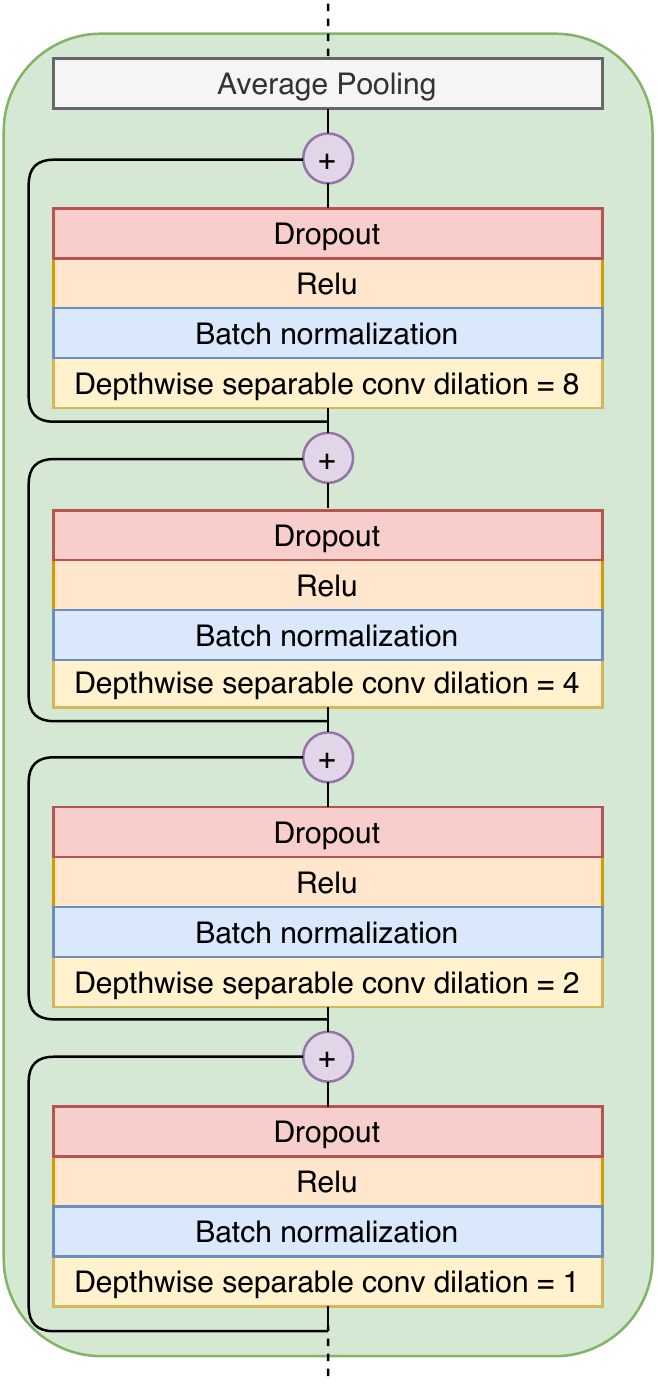}
        \caption{Base conv block\vspace{1em}}\label{fig:base_block} 
    \end{subfigure}
&
    \begin{subfigure}{0.25\linewidth}
        \centering
        \includegraphics[width=1.05\linewidth]{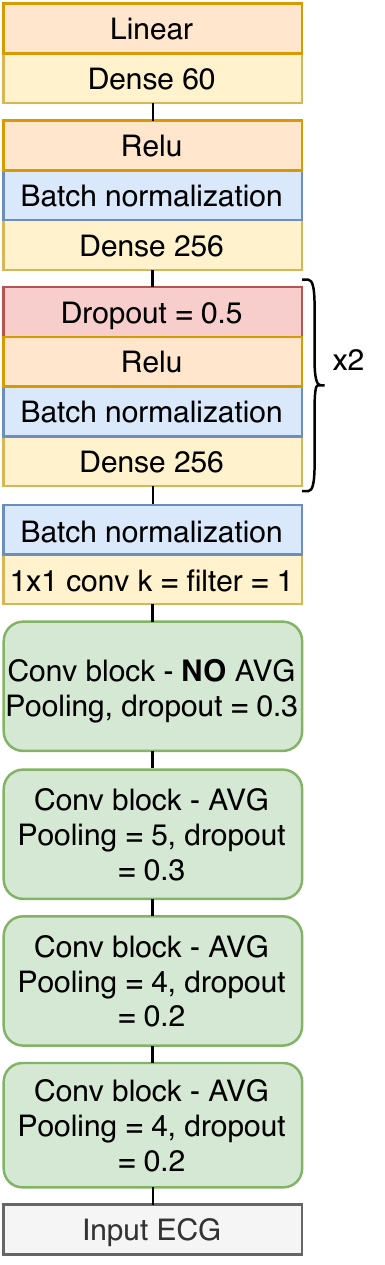}
        \caption{CNN + dense\vspace{1em}}\label{fig:cnn_dense} 
    \end{subfigure}
    \\ 
        \begin{subfigure}[t]{0.4\linewidth}
            \centering
            \includegraphics[width=0.52\linewidth]{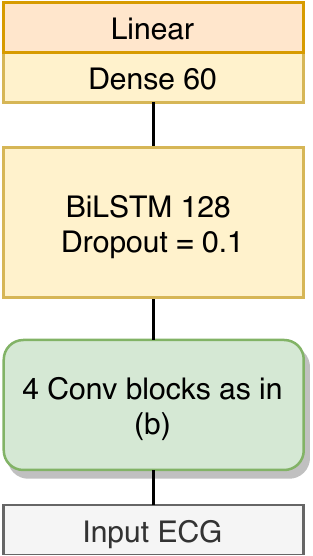}
            \caption{CNN + LSTM\vspace{-1.5em}}\label{fig:cnn_lstm}
        \end{subfigure}\vspace{2em} &
        \begin{subfigure}[t]{0.42\linewidth}
            \centering
            \includegraphics[width=\linewidth]{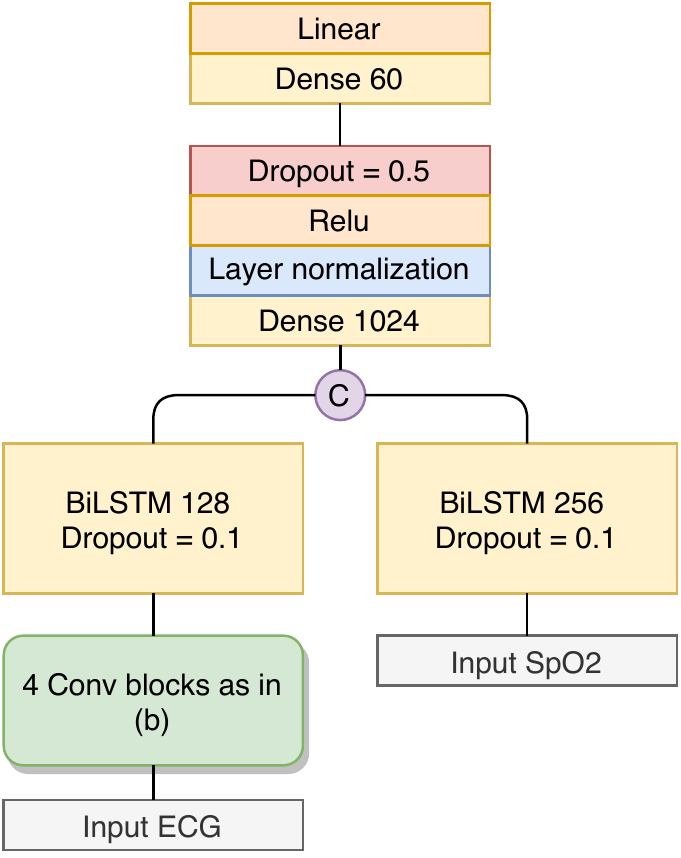}
            \caption{CNN + LSTM with SpO2\vspace{-1.5em}}\label{fig:cnn_lstm_sp}
        \end{subfigure}
\end{tabular}
    \caption{The considered deep learning architectures.}\label{fig:dnn_architectures}
\end{figure}

We performed two sets of experiments. The first set helped us in the development of the models, proving their worthiness against a well-recognized testbench. More precisely, we confronted our approach with several state-of-the-art solutions for automatic apnea detection by making use of the Apnea-ECG Database. In this case, a grid search-based hyperparameter tuning through 10-fold cross-validation on the training instances (each instance corresponding to an interval to be tagged) was carried out, while the final evaluation was conducted following the approaches adopted by the considered contenders.

The second set of experiments aimed at establishing the performance of the proposed models on the real-world Stroke Unit dataset described in Section \ref{sec:domain}. To perform hyperparameter tuning, the 30 patients were randomly partitioned into 2 disjoint subsets, making sure not to fragment the data belonging to each individual: 23 in the training set and 7 in the validation (tuning) set (see Table \ref{tab:pats} for details). The models were ultimately evaluated relying on leave-one-out cross-validation (each test fold corresponding to one of the 30 patients) and the results were also compared to those obtained from a 1D-ResNet architecture \cite{hong2019combining} on ECG, and an LSTM on SpO2 data, to provide a baseline.

Let $TP$ and $TN$ be respectively the instances (corresponding to 60-second intervals in Apnea-ECG Database, and single seconds in our dataset) correctly identified as apnea and normal breathing behaviour, $FP$ be the normal breathing instances erroneously recognized as apnea, and $FN$ be the apnea instances tagged as normal breathing. 
The considered evaluation metrics include
\smallskip
\begin{align}
accuracy = \frac{TP + TN}{TP + TN + FP + FN} \; ,
\end{align}
that provides a general assessment of the performance of the models concerning their capability of discerning between apnea and normal breathing instances. 

Since especially our dataset is severely unbalanced (there are way fewer seconds affected by apnea than those exhibiting a normal breathing behaviour), such a metric alone is not enough to correctly evaluate the models. 

To overcome its limitations, we also consider the following additional metrics:
\begin{align}
    sensitivity \ (or \ recall) = \frac{TP}{TP + FN} \; , 
\end{align}
that measures the capability of a model to identify apnea instances (models with high sensitivity are able of identifying a large fraction of the total apnea events);
\begin{align}
    specificity = \frac{TN}{TN + FP} \; ,
\end{align}
that measures the proportion of correctly identified normal breathing instances (models with a high specificity do not mistake normal breathing behaviour for apnea);
\begin{align}
F1 = \frac{2 TP}{2TP + FP + FN} \; ,
\end{align}
that provides a single score that measures the ability of the models of identifying all and only apnea instances as such.

In addition, we take into account the area under the ROC curve (AUC), which can be computed by comparing  the arrays of predicted anomalies with the ground truth values. Intuitively, it corresponds to the probability that a classifier will rank a randomly chosen positive (apnea) instance higher than a randomly chosen negative (normal breathing) one, assuming that \lq positive\rq{} ranks higher than \lq negative\rq . Such a metric is commonly used in the literature for model comparison purposes \cite{fawcett2006introduction}.

Only for our dataset, and limited to the best model, we also considered the following parameter:
\begin{align}
    precision = \frac{TP}{TP + FP} \; ,
\end{align}
that tracks the fraction of actual apnea instances among those identified as such by the model (note that the F1 score can be defined as the harmonic mean of precision and recall). 

Finally, we performed a per-patient evaluation, assessing the concordance between predicted and actual OSAS severity classes, and between OSA/non-OSA subjects ($\text{AHI} \geq 5$ threshold) based on the extracted AHI scores.

\subsection{The Neural Network Architecture}

The proposed architecture aims at providing a framework to handle ECGs, and possibly other physiological or waveform-like signals, in a straightforward and easily repurposable way. From a general perspective,
given an $s$ second-long signal sampled at $h$ Hz (e.g., the preprocessed ECG signal of Section~\ref{sec:data_preproc}), we derive a compact representation of size $k \times s$, where $k$ is the number of features that we want to extract each second, on which to perform the desired (classification) task. 

Specifically, we make use of a \mbox{1-D} convolutional neural network, exploiting depth-wise separable convolutions with dilation. 
As shown in Figure~\ref{fig:dnn_architectures}, the key components of the network are a set of arbitrarily stacked \emph{convolutional blocks}, each one characterized by a fixed series of operations, repeated four times (Figure \ref{fig:base_block}): $(i)$ depth-wise separable convolution with dilation; $(ii)$  batch normalization; $(iii)$ Relu activation function, and $(iv)$ spatial dropout. Between a series and the next one, a skip connection is employed to provide an alternative path for the gradient and avoid vanishing issues. At the end of each convolutional block, an average pooling operation is applied to reduce the size of the data. The block structure, e.g., the number of depthwise convolutional layers and their dilation, was empirically determined evaluating its performance on the validation sets. 
As for the arrangement (number and pooling sizes) of the stacked convolutional blocks, it has to be chosen so that the input signal is downsampled to 1-second granularity. Since several such configurations are possible, the best one was again determined by tuning on the validation sets.

After the convolutional blocks, data can be transformed in different ways according to the desired neural network architecture. As an example, if dense layers are to be put after the convolutional blocks, data can pass through a 1x1 convolution (Figure \ref{fig:cnn_dense}). In this case, starting from the 14400 input ECG values of the Stroke Unit dataset, we end up with 180 values (due to the chosen pooling sizes), which intuitively encode the condensed 1-second granularity representation of the original information. Another possibility (Figure \ref{fig:cnn_lstm}) is that of stacking an LSTM directly on the convolutional output, which, in our case, can be seen as a multivariate, 180-second long time series, and then considering the last output of the sequence. 
In both variants, the depicted neural networks generate, as the final outcome, 60 values, each representing a non-thresholded score related to the likelihood of having an apnea in each considered second. It is worth pointing out that there is no activation function in the output layer, because of the choice of weighted squared hinge loss function: an anomaly will be considered as present when the output is greater than 0, absent otherwise. This is the setting employed for our dataset. In the case of Apnea-ECG, just a single score is returned, due to the 1-minute granularity classification task. To train the model, we exploited Adam optimizer, one-cycle learning rate scheduler, and gradient clipping. 

\begin{figure*}[tb]
    \centering
    \includegraphics[width=\linewidth]{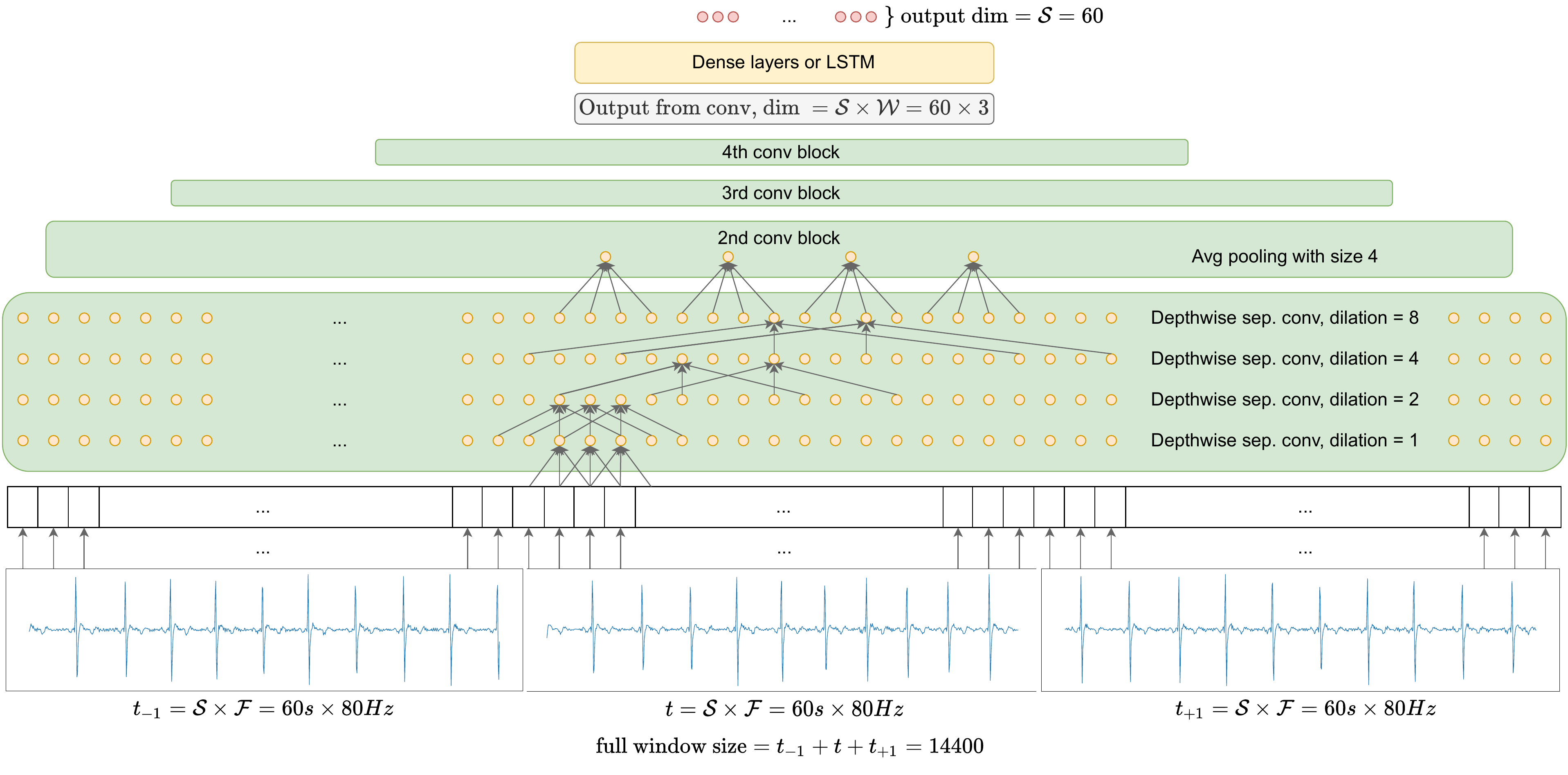}
    \caption{Hi-level representation of the AIOSA model showing how temporal resolution is reduced during the computation. $\mathcal{S}$ is the single window size in seconds, $\mathcal{F}$ is the sampling frequency in Hz, and $\mathcal{W}$ is the number of considered windows (3 in the example).} 
    \label{fig:temporal_model}
\end{figure*}

A graphical account of the whole process is given in Figure \ref{fig:temporal_model}. It provides an intuitive hi-level representation of how an AIOSA model reduces the temporal resolution of input data to 1-second granularity by exploiting its distinctive pyramidal structure.

We  conclude the section by explaining the most relevant choices that led us to the development of the architecture. Depth-wise separable convolutions allow us to save computation with respect to classic convolutions, as the network is able to process more in a shorter amount of time; a kernel size of 3 and a number of filters equal to 16 proved to empirically fit our case. Dilations are employed to look at varying (and wider) temporal frames of input data. We prefer average with respect to max-pooling as it should behave better at preserving localization (temporal information in our case).  In addition,  average pooling has empirically performed better than max pooling on preliminary tests. Last but not least, by adjusting the pooling values, we can deal with different signal sampling rates. As for the loss function, we chose weighted squared hinge over classic binary cross-entropy to maximize the margin between predictions, taking into account that in the dataset there are far more seconds characterized by normal breathing than those affected by anomalies.  As for the optimizer setting, one-cycle learning rate scheduler has been observed to favour model generalizability, while gradient clipping is typically employed with LSTMs to avoid exploding gradient issues. 
As for the specific optimizer hyperparameters, we refer the reader to Table~\ref{tab:opt} (others are left at their default values).

\begin{table}[tb]
\centering
\caption{Optimizer hyperparameters.\label{tab:opt}}
\resizebox{\linewidth}{!}{%
\begin{tabular}{ccccccc}
\toprule
\multirow{2}{*}{\centering Dataset} & \multirow{2}{*}{\centering Base LR} & \multirow{2}{*}{\centering Max LR} & \multirow{2}{*}{\centering Epochs} & \multirow{2}{5em}{\centering Max Epochs} & \multirow{2}{5em}{\centering Base momentum} & \multirow{2}{5em}{\centering Max momentum} \\\\
\midrule 
Apnea-ECG & 3e-4 & 1e-1 & 239 & 300 & 0.85 & 0.95 \\
Our & 3e-4 & 1e-2 & 191 & 200 & 0.78 & 0.99\\
\bottomrule
\end{tabular}
}
\end{table}

\subsection{A Note on SpO2 Data}

As witnessed by the literature \cite{wali2020correlation}, oxygen saturation indexes are strongly correlated with OSA severity. 
For this reason, although we mainly focus is on routinely monitored ECG data, we also considered SpO2 under two different perspectives on our Stroke Unit dataset. 

First, as a standalone piece of information, encoded as a 1-second granularity time series, given as input to a vanilla bidirectional LSTM; this is useful to provide us with a baseline. Similarly to the ECG model, the recurrent neural network relies on 180-second windows to predict a list of 60 binary values. 

A second, more advanced approach exploits neural network compositionality by merging the latent representation obtained as output from the bidirectional LSTM on SpO2 data with the one provided by the CNN-LSTM model on ECG data (Figure~\ref{fig:cnn_lstm_sp}). The idea is to combine two temporal-aware representations referring to the same time interval, but based on different, possibly complementary, predictors.

As a final note, we would like to point out that, as mentioned in Section \ref{sec:intro},  SpO2 can also be considered a routinely recorded signal, being usually provided by a fingertip photoplethysmogram.

\section{Results}
\label{sec:res}

In this section, the outcomes of the evaluation of the proposed models on both the Apnea-ECG database and our Stroke Unit dataset are reported.

\subsection{Results on the Apnea-ECG database}

\begin{table*}[t]
\centering
\caption{Performance on the Apnea-ECG database.\label{tab:confrapnecg}}
\resizebox{\linewidth}{!}{%
\begin{tabular}{cccccccccccc}
\toprule
\multirow{2}{*}{\centering Paper}  & \multirow{2}{*}{\centering Method}  & \multicolumn{5}{c}{Per-segment}  & \multicolumn{4}{c}{Per-patient {\footnotesize(OSA = AHI $\geq$ 5)}}     \\      
&                 & Sens  & Spec  & Acc   & F1 & AUC &    Sens  & Spec  & Acc   & F1 &    \\ \midrule
\cite{DBLP:conf/cinc/SadrC14} & \multirow{4}{*}{\begin{tabular}[c]{@{}c@{}}training \\ and test\end{tabular}}     & 0.813 & 0.917 & 0.877 & - & -  & -  & -  & -  &  -  \\
\cite{DBLP:journals/cbm/SharmaS16}                 &       & 0.795  & 0.884  & 0.838  & -  & 0.834  &  0.958 &  \textbf{1.0} & 0.971  &   -  \\
\cite{DBLP:journals/tbe/SongLZCX16}                 &       & 0.826  & 0.884  & 0.862  & -  & 0.940  & 0.958  & \textbf{1.0}  & 0.971  & -     \\
\cite{li2018method}                 &       & 0.889  & 0.821  & 0.847  & -  & 0.869  & \textbf{1.0}  & \textbf{1.0}  & \textbf{1.0}  & \textbf{1.0}     \\
\cite{singh2019novel}                 &       & 0.900  & 0.838  & 0.862  & -  & 0.881  & \textbf{1.0}  & \textbf{1.0}  & \textbf{1.0}  & \textbf{1.0}    \\
\cite{DBLP:journals/sensors/ChangYLL20}                 &       & 0.811  & 0.920  & 0.879  & 0.865  & 0.935  & 0.957  & \textbf{1.0}  & 0.971  & -     \\
\cite{sharma2020sleep}                 &       & 0.849 & 0.882 & 0.875 & - & 0.930  & \textbf{1.0}  & 0.909  & 0.971  & -    \\
\cite{feng2020sleep}                 &       & 0.862  & 0.844  & 0.851  & 0.814  & -  & 0.957  & \textbf{1.0}  & 0.971  & -      \\
\cite{shen2021multiscale}                 &       & 0.898  & 0.891  & 0.894  & 0.866  & 0.946  & \textbf{1.0}  & \textbf{1.0}  & \textbf{1.0}  & \textbf{1.0}    \\
\cite{martin2017heart}                 &       & 0.814  & 0.868  & 0.848  & -  & 0.920  & 0.95  & \textbf{1.0}  & 0.967  & -    \\
\cite{wang2019sleep}                 &       & 0.831  & 0.903  & 0.876  & -  & 0.950  & \textbf{1.0}  & 0.917  & 0.971  & -     \\
Our (CNN + dense)                       &                                                                                    & 0.907 & 0.937 & 0.925 & 0.902 & 0.976  & \textbf{1.0}  & 0.917  & 0.971 & 0.979       \\
Our (CNN + LSTM)                        &                                                                                  & \textbf{0.912} & \textbf{0.951} & \textbf{0.936} & \textbf{0.916} & \textbf{0.981}   & \textbf{1.0}  & \textbf{1.0}  &    \textbf{1.0} & \textbf{1.0} \\ \midrule
\cite{fatimah2020detection}             & \multirow{4}{*}{\begin{tabular}[c]{@{}c@{}}10-fold CV\\  on training\end{tabular}} & 0.897  & \textbf{0.947} & 0.926 & 0.905 & 0.970  & -  & -  & -  &  -    \\
\cite{almutairi2020detection}          &                                                                                    & 0.899 & 0.879 & 0.891 & 0.914 & -   & -  & -  & -  &  -   \\
Our (CNN + dense)                       &                                                                                    & 0.925 & 0.927 & 0.926 & 0.905 & 0.977 & -   & -  & -  & -     \\
Our (CNN + LSTM)                        &                                                                                    & \textbf{0.951}     & 0.937     & \textbf{0.943}     & \textbf{0.927}     & \textbf{0.987}  & -   & -  & -  & -     \\ \bottomrule
\end{tabular}%
}
\end{table*}

Table \ref{tab:confrapnecg} shows the performance of the models on the Apnea-ECG database. We considered two architectural variants that differ from each other in their final components: the first one is based on 1x1 convolution layers followed by dense layers (Figure \ref{fig:cnn_dense}); the other one makes use of a 16-features LSTM (Figure \ref{fig:cnn_lstm}), as the size of the output of the convolutional blocks is $180 \times 16$. Of course, in this case, there is a single output neuron, instead of 60, as the goal was that of labeling single minutes with the presence or absence of apnea. Another difference with respect to the depicted architectures is that the pooling values are 4-5-5, instead of 4-4-5, as in the dataset ECG information is sampled at 100 Hz, instead of 80 Hz. The obtained results are compared with those given in \cite{DBLP:conf/cinc/SadrC14,DBLP:journals/cbm/SharmaS16,DBLP:journals/tbe/SongLZCX16,li2018method,singh2019novel,DBLP:journals/sensors/ChangYLL20,almutairi2020detection,sharma2020sleep,feng2020sleep,shen2021multiscale,martin2017heart,wang2019sleep,fatimah2020detection}. We focus on these contributions as they are very recent, provide two different evaluation methodologies (10-fold cross-validation on training data, and official training + test split), and outperform previous state-of-the-art approaches. In most cases \cite{DBLP:conf/cinc/SadrC14,DBLP:journals/cbm/SharmaS16,DBLP:journals/tbe/SongLZCX16,li2018method,singh2019novel,DBLP:journals/sensors/ChangYLL20,sharma2020sleep,feng2020sleep,shen2021multiscale,martin2017heart,wang2019sleep}, the model was developed taking into account the 35 training set instances, and then evaluated on the official 35 test set instances. In \cite{almutairi2020detection,fatimah2020detection}, both model development and evaluation were performed via 10-fold cross-validation over the training set, where each instance corresponds to an interval to be tagged (even though not the best method). 

By applying the same scoring methods, we showed that both our models vastly outperform previous proposals, with the overall best results provided by the LSTM-based variant, considering both per-segment and per-patient results. More precisely, as for per-segment results, an accuracy improvement of 8.3\% and 4.7\% was obtained with respect to, respectively, the average and best performance of the considered state-of-the-art approaches. As for the OSA/non-OSA patient detection, our model provides a perfect classification result. 
This is quite interesting, as it supports the intuition that the proposed convolutional architecture is able to effectively summarize input information while preserving its temporal content.

\subsection{Results on the Stroke Unit dataset}

\begin{table*}[t]
\centering
\caption{Performance comparison on the Stroke Unit dataset considering leave-one-patient-out CV.\label{tab:confrour}}
\resizebox{\linewidth}{!}{%
\begin{tabular}{cccccccccccccc}
\toprule
\multirow{2}{*}{\centering Method} & \multirow{2}{*}{\centering Model} & \multirow{2}{*}{\centering Signal}  & \multicolumn{5}{c}{Per second}  & \multicolumn{4}{c}{Per-patient {\scriptsize(OSA = AHI $\geq$ 5)}}   & \multicolumn{2}{c}{Per-patient {\scriptsize(AHI class)}  }  \\      
 & &       &     Sens  & Spec  & Acc   & F1 & AUC &   Sens  & Spec  & Acc  &  F1 & Acc@1 & Acc@2  \\
  \midrule
\multirow{4}{5em}{\centering All patients} & ResNet & ECG & 0.168 & 0.824  & 0.716 & 0.162 & 0.523 & 0.087 & \textbf{0.857}  & 0.267  & 0.154  & 0.233  &  0.433     \\
& LSTM & SpO2 & 0.656 & 0.680 & 0.676 & 0.399 & 0.704 & 0.826 & 0.0  & 0.633 &  0.776  & 0.300  &  0.833        \\
& CNN + LSTM & ECG & 0.628 & 0.796 & 0.769 & 0.471 & 0.750 & 0.870 & 0.286  & 0.733  & 0.833   &  0.433 &  0.833     \\
& CNN + LSTM & ECG + SpO2 & \textbf{0.672} & \textbf{0.843} & \textbf{0.815} & \textbf{0.543} & \textbf{0.825} &  \textbf{1.0}  & 0.714  & \textbf{0.933}  & \textbf{0.958} & \textbf{0.666}  & \textbf{1.0}         \\
 \midrule
\multirow{4}{5em}{\centering Without validation patients} & ResNet & ECG & 0.107 & \textbf{0.865}  & 0.737 & 0.121 & 0.525 & 0.111 & \textbf{0.800}  & 0.261  & 0.190  & 0.217  &  0.391     \\
& LSTM & SpO2 & 0.624 & 0.698 & 0.685 & 0.402 & 0.704 & 0.833 & 0.0  & 0.652 &  0.789  & 0.261  &  0.783        \\
& CNN + LSTM & ECG & 0.643 & 0.774 & 0.752 & 0.468 & 0.760 & 0.889 & 0.200  & 0.739  & 0.842   &  0.435 &  0.782     \\
& CNN + LSTM & ECG + SpO2 & \textbf{0.655} & 0.849 & \textbf{0.816} & \textbf{0.547} & \textbf{0.826} &  \textbf{1.0}  & 0.600  & \textbf{0.913}  & \textbf{0.947} & \textbf{0.609}  & \textbf{1.0}      \\
\bottomrule
\end{tabular}
}
\end{table*}

\begin{figure*}[tb]
	\centering
\begin{subfigure}{\textwidth}
    \centering
    \includegraphics[width=\linewidth,]{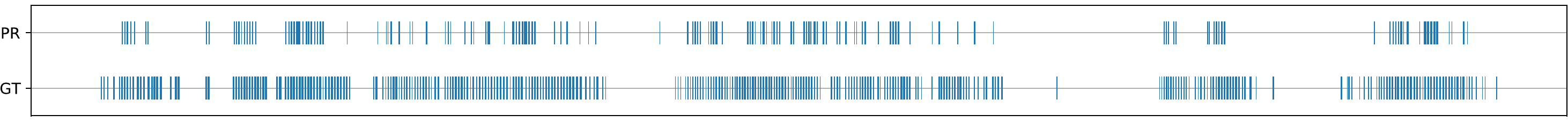}
    \caption{Patient 15}\label{fig:barre_15} 
\end{subfigure}
\\
\medskip
\begin{subfigure}{\textwidth}
    \centering
    \includegraphics[width=\linewidth,]{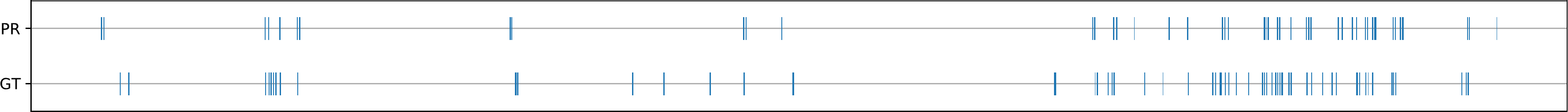}
    \caption{Patient 19}\label{fig:barre_19} 
\end{subfigure}
\\
\medskip
\begin{subfigure}{\textwidth}
    \centering
    \includegraphics[width=\linewidth,]{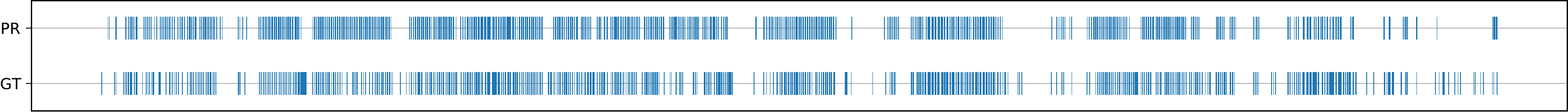}
    \caption{Patient 26}\label{fig:barre_26} 
\end{subfigure}
	\caption{Predicted (PR) and ground truth (GT) events for three patients.}
	\label{fig:barre}
\end{figure*}

The evaluation on the Stroke Unit dataset is based on leave-one-out cross-validation, where each test fold corresponds to one of the 30 patients. 
Given that CNN-LSTM architectures had shown remarkable performance on the Apnea-ECG database, we considered two variants of them. The first one, depicted in Figure \ref{fig:cnn_lstm}, makes use of just ECG data; the second one, shown in Figure \ref{fig:cnn_lstm_sp}, also relies on the SpO2 signal. 
For the sake of comparison, a classic 1-D ResNet \cite{hong2019combining} was tested on ECG data, and a vanilla bidirectional LSTM was applied to SpO2 (both of them were tuned according to the same training/validation split as of our models). 

Table \ref{tab:confrour} reports the achieved performance concerning both the per-second and per-patient classification. 
As for the per-second results, they were micro-averaged across the folds. ResNet showed the worst behaviour, not being able to extract valuable information from the ECG data. LSTM had a better and encouraging outcome, suggesting that SpO2 is indeed a good predictor, despite the large amount of missing data, as domain experts confirm. Overall, the best figures were provided by our CNN+LSTM architecture, which excels when combined with SpO2, achieving an improvement of 57.7\%, 17.2\%, and 10\% compared to respectively ResNet, LSTM, and CNN+LSTM itself when applied on only ECG data. 
Results are in line with those reported by the extended study on human-human and human-machine sleep annotation concordance in \cite{DBLP:conf/embc/ThoreyHAD19}, where more stringent patient exclusion criteria and far richer PSG data were considered. 

Turning to per-patient results, in order to adhere to the literature on OSAS, where only anomalous respiratory events lasting at least 10 seconds are considered, we performed a light post-processing of the output, so to discard those blocks of anomalies that span less than 10 seconds, still preserving the F1 score. To this end, we set up the following optimization process,  based again on the validation set: $(i)$ we transformed the raw scores to probability-like values by applying a sigmoid function, and then $(ii)$ we looked for the best rounding threshold (0.5875) which, combined with the filtering procedure, provides the best F1 score. Once more, the best performance is exhibited by the CNN+LSTM architecture on ECG + SpO2, both for the task of discerning between OSA and non-OSA patients (with an AHI threshold of 5), and for the task of identifying the patients' AHI classes. The proposed model is able to almost perfectly detecting OSA/non-OSA patients with an F1 of 0.958 (0.947 disregarding validation set patients), while it reached an accuracy@1 of 0.666 and an accuracy@2 of 1.0 (respectively 0.609 and 1.0 disregarding validation set patients) on the AHI class detection task, meaning that, when an error is made, it is at most of a single class. 

Looking at the single subjects, the right-hand side of Table \ref{tab:pats} focuses on the per-patient results obtained by the CNN+LSTM architecture on ECG + SpO2 data evaluated in leave-one-out cross-validation. As it can be seen, among wrongly predicted ones, the majority (60\%) was considered as more serious than it actually was, which is a safe behaviour from the clinical point of view, while the others were just downgraded from \emph{severe} to \emph{moderate} class. This is a very good outcome, as the main goal for the physicians is that of identifying subjects who suffer from moderate to severe OSAS, in order to treat them accordingly.

As a final remark, we notice that, although F1 scores may seem low for some patients, they are sometimes explainable by the corresponding very high specificity values. This is the case, for instance, with Patient 18 and Patient 23, whose AHI is very low. 

In general, all metrics, but AHI, are penalized in our task. As an example, a correctly predicted, though slightly shifted, event would still be considered as an error. To support that, Figure \ref{fig:barre} shows predicted and ground truth anomalies for patients 15 (Figure \ref{fig:barre_15}), 19 (Figure \ref{fig:barre_19}), and 26 (Figure \ref{fig:barre_26}). Although patient 15 has an F1 of 0.446 and gets classified as \emph{moderate}, instead of \emph{severe}, the model is still able of identifying clusters of anomalies that may be relevant for the clinical decision-making process. Patient 19, despite showing an even lower (0.249) F1 score, is correctly classified, with a good approximation of his/her AHI and anomalies distribution. Finally, Patient 26, who is correctly classified as well, has a high (0.701) F1 score.

Figures like the above one, which are made possible by the adopted 1-second tagging strategy, are particularly useful as they provide clinicians with fine-grained information about the condition of a patient, and allow them to better interpret and validate the results of the model.

\subsection{Effects of Reducing the Training Data Size}

\begin{figure}[t]
    \centering
    \includegraphics[width=\linewidth]{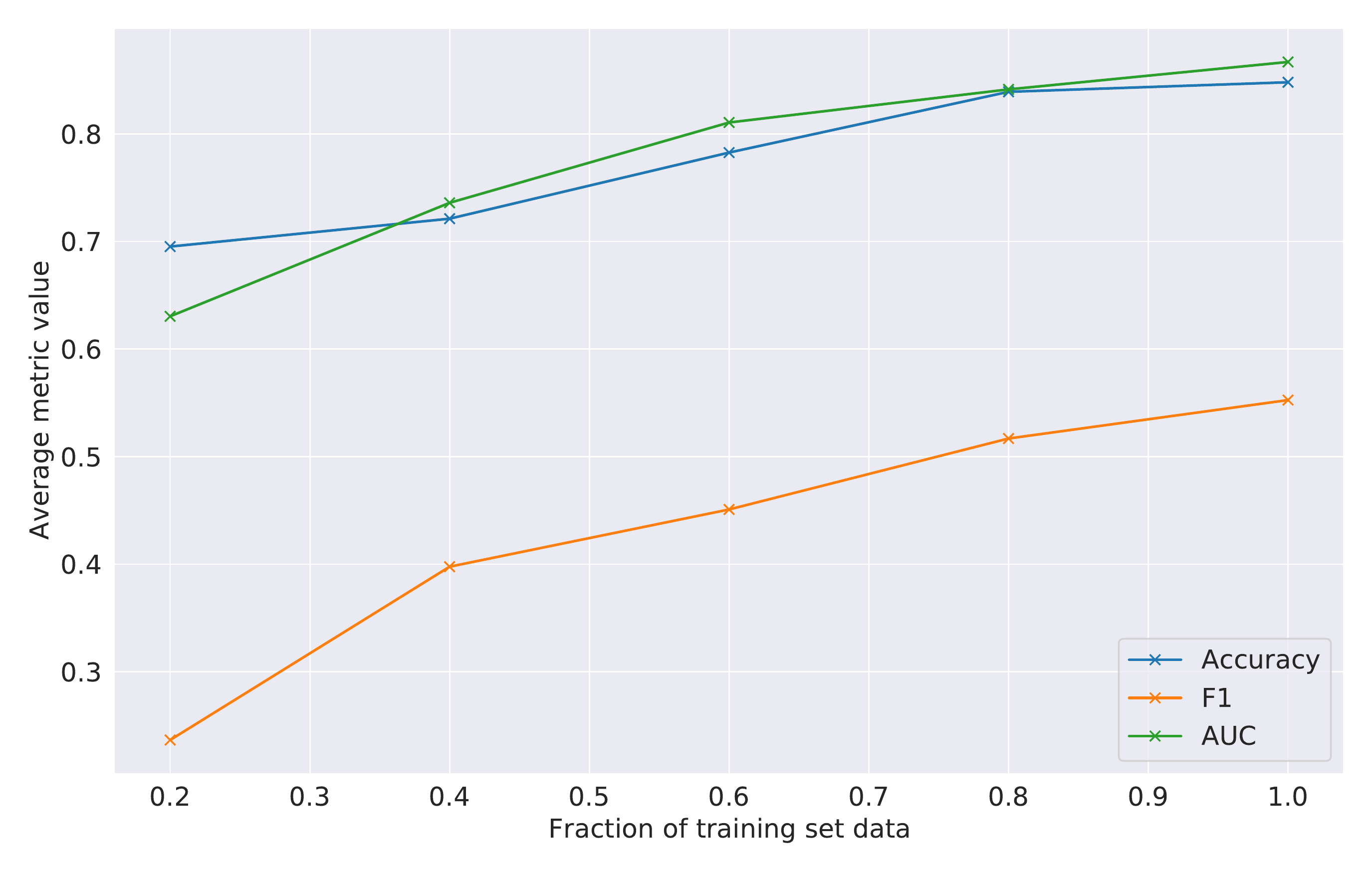}
    \caption{Best model results on the Stroke Unit dataset obtained by varying the training data size.}
    \label{fig:trainred}
\end{figure}

To conclude, let us analyze how the performance of the CNN+LSTM architecture with ECG + SpO2 data varies on the Stroke Unit dataset when increasingly larger parts of the training patients are discarded. 
To such an extent, we considered a fraction of 0.2, 0.4, 0.6, 0.8, and 1.0 of the original training set of patients for model learning purposes, while the results are always established against the same validation set of patients according to accuracy, F1, and area under the ROC curve. In order to obtain a reliable estimate of the chosen metrics, for each fraction we repeated the training and evaluation steps on 10 randomly sampled subsets of patients, averaging the final results. As it emerges from Figure \ref{fig:trainred}, all metrics improve as the size of the training set increases, meaning that higher performance of the model must be expected when new patients are included in the learning phase. Such a behaviour should not come as a surprise, given the heterogeneity and peculiarities of the patients belonging to our dataset. Nevertheless, note that, while a high gain in F1 score was observed when increasing the fraction of training data from 0.2 to 0.4, a law of diminishing returns seems to apply, especially after the 0.8 threshold.

\section{Conclusions and Future Work}

The ultimate goal of the present work was to develop an effective and easily deployable tool to help the clinical decision-making process in the context of OSAS. To this end,
we proposed a deep learning framework for the detection of sleep apnea events, based on convolutional neural networks. Its distinctive feature is that it is able to deal with waveform data, such as physiological signals, by effectively summarizing them and extracting their key components. We first applied the framework to the well-known Apnea-ECG database, and proved that it outperforms current state-of-the-art solutions. Then, we focused on a real and challenging scenario, namely, acute stroke patients admitted to the Clinical Neurology Unit of the Udine University Hospital. 

As already pointed out, the presence of severe OSAS in these patients is associated with higher mortality, worse neurological deficits, worse functional outcome after rehabilitation, and a higher likelihood of uncontrolled hypertension, and thus their early detection is of fundamental importance.
Unlike previous studies, which considered stringent exclusion criteria for both the patients and the quality of their recordings, our data are strongly affected by noise, and individuals may suffer from several comorbidities. 

Based on leave-one-out cross-validation, we showed that the proposed solution is able to correctly identifying OSAS cases in the dataset, and to assess their severity, based on routinely recorded vital signs, such as ECG and oxygen saturation, only. Moreover, the model provides physicians with fine-grained information about the condition of the patient, which is useful for explainability and validation purposes. This is an extremely encouraging achievement, especially considering that, as it emerged from the in-depth review of the literature, such a task, when conducted on real-world data and without polysomnography equipment, turns out to be quite hard for AI systems, and impossible for a human scorer.

As for future work, the achieved results and how they have been obtained show the great flexibility of the proposed architecture. We plan to investigate its use as a module of larger models on different datasets and/or tasks characterized by waveform-like data.
We also intend to experiment with Conditional Random Fields, as the single outputs of the network have a clear correlation (seconds marked as apneas tend to cluster together).
Moreover, a model may be developed that, taking the results of the current one as input, is able to classify the predicted anomaly as central apnea, obstructive apnea, mixed apnea, or hypopnea.
The outcomes of the experimentation also highlighted the opportunity of investigating  possible refinements of the considered metrics, in order to make them more suitable to performance evaluation in the considered use case.
The final, natural development of the work is the integration of the model in a production setting, as a support for the hospital's activities.

\section*{Acknowledgements}
This work was supported by Google Academic Research Grant and TensorFlow Research Cloud programs, as well as the Italian INdAM-GNCS project \emph{Ragionamento Strategico e Sintesi Automatica di Sistemi Multi-Agente}. Moreover, the authors would like to thank the reviewers for their valuable comments and suggestions, which helped them in improving the paper.

\bibliography{mybibfile}

\end{document}